%% file: main.tex
\documentclass[runningheads]{llncs}

\usepackage{eccv}

\usepackage{eccvabbrv}
\usepackage{booktabs}
\usepackage{array}
\usepackage{graphicx}
\usepackage{multirow}
\usepackage{pifont}
\usepackage{graphicx}
\usepackage[accsupp]{axessibility}  
\usepackage{adjustbox}
\usepackage{booktabs,tabularx,array,xcolor,makecell,colortbl}
\definecolor{cardgray}{RGB}{245,246,250}
\definecolor{sectiongray}{RGB}{235,237,242}

\usepackage{makecell}
\usepackage[table]{xcolor}

\newcommand{\cmark}{\ding{51}}
\newcommand{\xmark}{\ding{55}}
\newcommand{\pmark}{\ensuremath{\sim}}

\usepackage[table]{xcolor}
\usepackage[most]{tcolorbox}
\usepackage{xcolor}
\definecolor{myblue}{RGB}{31, 119, 180}
\definecolor{myred}{RGB}{214, 39, 40}
\definecolor{mygreen}{RGB}{44, 160, 44}
\definecolor{bestcell}{RGB}{188,239,196}
\definecolor{secondcell}{RGB}{219,242,224}
\newcommand{\best}[1]{\cellcolor{bestcell}{#1}}
\newcommand{\second}[1]{\cellcolor{secondcell}{#1}}
\usepackage[table]{xcolor}
\newtcolorbox{hypobox}{
  colback=blue!5!white,
  colframe=blue!40!black,
  boxrule=0.6pt,
  arc=2mm, 
  left=6pt, right=6pt, top=4pt, bottom=4pt,
  before skip=10pt, after skip=10pt
}
\newtcolorbox{conbox}{
  colback=green!5!white,
  colframe=green!40!black,
  boxrule=0.6pt,
  arc=2mm,
  left=6pt, right=6pt, top=4pt, bottom=4pt,
  before skip=10pt, after skip=10pt
}
\newtcolorbox{quesbox}{
  colback=orange!8!white,
  colframe=orange!60!black,
  boxrule=0.6pt,
  arc=2mm,
  left=6pt, right=6pt, top=4pt, bottom=4pt,
  before skip=10pt, after skip=10pt
}
\newtcolorbox{redbox}{
  colback=gray!10!white,
  colframe=red!60!black,
  boxrule=0.6pt,
  arc=2mm,
  left=6pt, right=6pt, top=3pt, bottom=3pt,
  before skip=10pt, after skip=10pt
}
\newtcolorbox{greenbox}{
  colback=gray!10!white,
  colframe=green!60!black,
  boxrule=0.6pt,
  arc=2mm,
  left=6pt, right=6pt, top=3pt, bottom=3pt,
  before skip=10pt, after skip=10pt
}
\newtcolorbox{bluebox}{
  colback=gray!10!white,
  colframe=blue!60!black,
  boxrule=0.6pt,
  arc=2mm,
  left=6pt, right=6pt, top=3pt, bottom=3pt,
  before skip=10pt, after skip=10pt
}

\usepackage{hyperref}

\usepackage{orcidlink}

\begin{document}

\title{\textit{WildCity}: A Real-World City-Scale Testbed for Rendering, Simulation, and Spatial Intelligence} 

\titlerunning{WildCity}

\author{Xiangyu Han\inst{1,2} \and
Mengyu Yang\inst{2} \and
Jiaqi Li\inst{2} \and
Bowen Chang\inst{2} \and
Ziyu Chen\inst{4} \and \\
Hexu Zhao\inst{2} \and
Rahul Kumar Agrawal\inst{1} \and
Anthony Rodriguez\inst{1} \and
Rajani Acharya\inst{1} \and
Fiona Hua\inst{1} \and
Marco Pavone\inst{3,4} \and
Chen Feng\inst{2}$^\dagger$ \and
Yiming Li\inst{2,3}$^\dagger$
}

\authorrunning{Han et al.}

\institute{
May Mobility, Ann Arbor, MI, USA \and
New York University, New York, NY, USA \and
NVIDIA, Santa Clara, CA, USA \and
Stanford University, Stanford, CA, USA
\url{https://han-xiangyu.github.io/Wild-City/}}
\vspace{-0.3cm}

\maketitle
\vspace{-3pt}
\begin{center}
    \centering
    \captionsetup{type=figure}
    \includegraphics[width=\textwidth]{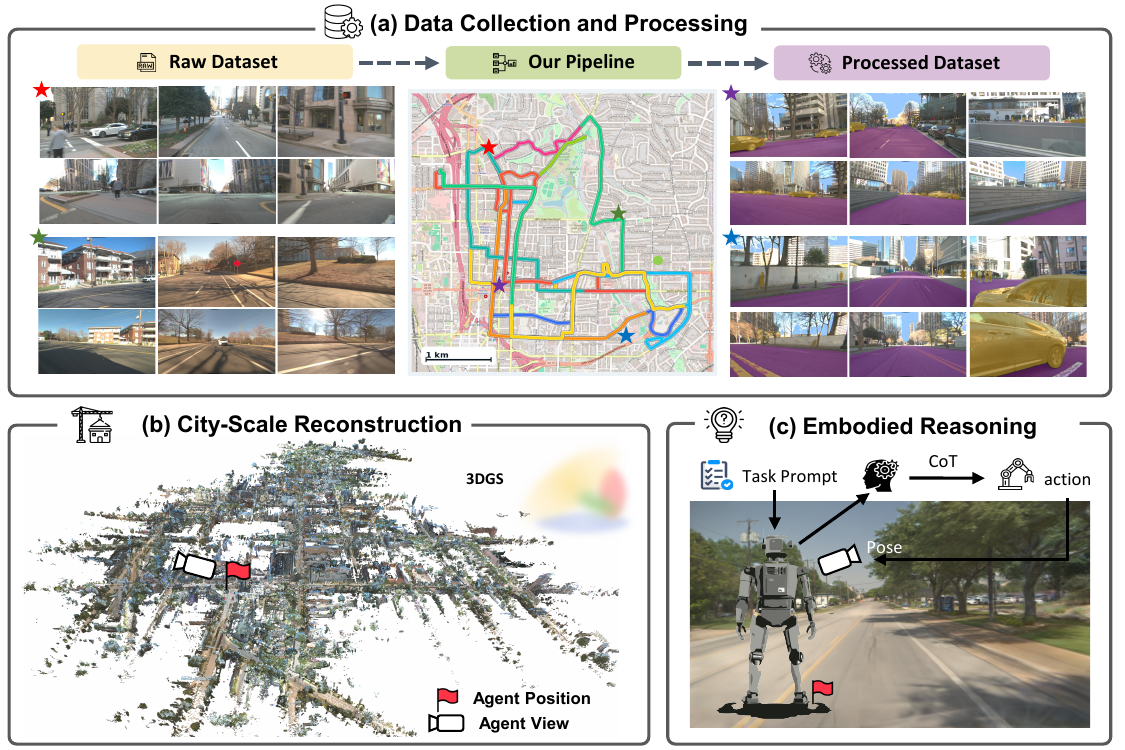}
    \vspace{-5pt}
    \captionsetup{width=\textwidth}
    \caption{\textbf{Overview of \textit{WildCity}.} It contains 18 long-horizon trajectories over 1,500\,km across six real-world cities. We process raw autonomous-fleet logs into multimodal data for city-scale reconstruction, simulation, and closed-loop embodied reasoning.}
    \vspace{-5pt}
    \label{fig:teaser}
\end{center}

\begingroup
\renewcommand{\thefootnote}{\fnsymbol{footnote}}
\footnotetext[4]{Corresponding authors.}
\endgroup
 
\input{sec/0_abstract}  
\input{sec/1_intro}
\input{sec/2_related_works}

\input{sec/3_wildcity_dataset}

\input{sec/4_wildcity_baseline}

\input{sec/5_experiments}
\input{sec/6_discussion}

\input{sec/7_conclusion}

\section*{Acknowledgements}
We thank the May Mobility operations, data platform, and autonomy teams for their support in real-world data collection, on-site operations, and data processing infrastructure. Chen Feng was supported by NSF grant No.~2238968.

%
%

\clearpage
\bibliographystyle{splncs04}
\bibliography{main}

\clearpage
\input{sec/X_suppl}

\end{document}

%% file: sec/0_abstract.tex
\begin{abstract}
Humans can navigate an unfamiliar city and gradually form a coherent spatial mental map spanning tens of square kilometers. Can AI build spatial representations at a comparable scale? Although recent foundation models have advanced scene reconstruction and embodied intelligence, scaling to entire cities remains an open challenge, primarily due to the lack of city-scale data. To bridge the gap, we introduce \textbf{WildCity}, a real-world multimodal dataset collected by autonomous fleets traversing complex urban environments. Our dataset includes \textbf{18 trajectories}, each averaging \textbf{83.7 kilometers} in length, and preserves the core challenges of in-the-wild perception, \eg, dynamic objects, lighting variations, and imperfect camera poses. We further establish an urban-tailored reconstruction baseline and convert the reconstructed environments into a closed-loop simulator. Beyond the dataset and baseline, we systematically analyze the key challenges on the path to simulation-ready urban digital twins: \textbf{scalability}, \textbf{extrapolation}, and \textbf{uncertainty}. Ultimately, WildCity aims to catalyze progress not only in city-scale rendering, but more broadly in the pursuit of AI that can perceive, remember, and reason across space at a scale comparable to human cognition.
    \keywords{Neural Rendering  \and  Dataset and Benchmark  \and Spatial AI}
\end{abstract}

%% file: sec/1_intro.tex
\section{Introduction}
\label{sec:intro}

Humans can wander for hours through the streets of a city—turning down narrow alleys, crossing wide plazas, catching glimpses of a distant tower between buildings—and still, over time, piece together a coherent \textit{spatial mental map} spanning tens of square kilometers. Such an internal representation of large-scale space can enable self-localization, long-term memory, and efficient planning~\cite{epstein2017cognitive,bellmund2018navigating, maguire2000navigation}. This leads to a question: \textit{Can AI, like a human, internalize the structure of an entire city from visual observations and use that knowledge to reason, plan, and act?}

Recent advances in foundation models, \eg, vision-language models (VLMs), have improved embodied intelligence across both virtual and real-world spatial environments—enabling tasks such as visual navigation~\cite{Khanna_2024_goat}, vision-language navigation~\cite{zheng2024navillm}, and active situated reasoning~\cite{yu2025thinking360deghumanoidvisual}. Yet these capabilities are typically demonstrated in small-scale scenarios—a single room, a synthetic apartment, or a city block. When scaling up spatial coverage or video duration, current models struggle to maintain spatial coherence and reason over long-range dependencies~\cite{yang2025cambrian}. \textit{This reveals a critical gap between the effortless scalability of human spatial cognition and the brittle, small-world reasoning of foundation models.}

To bridge this gap, foundation models require access to continuous, long-range visual-spatial observations that capture both the scale and the complexity of real-world environments. Yet existing datasets fall short. Synthetic data offers controllability but suffers from a substantial sim-to-real gap~\cite{li2023matrixcity}, while real-world benchmarks are typically limited to short video clips of isolated urban scenes~\cite{caesar2020nuscenes,sun2020scalability,li2024multiagent}. Without large-scale real-world visual-spatial data, it remains infeasible to build photorealistic city simulators—digital environments that faithfully mirror the complexity of real cities—and, in turn, to use such simulators for training and evaluating embodied agents powered by foundation models at scale.

In this work, we introduce \textbf{WildCity}, a real-world dataset and testbed for city-scale spatial intelligence. Collected by autonomous vehicle fleets traversing complex urban environments, WildCity provides continuous, surround-view multimodal data with city-scale spatial coverage (\textit{over 1,500 km of traversed roads}), diverse urban scenes (\textit{covering distinct functional zones}), and long sensory streams (\textit{averaging 2.5 hours per log}). In addition to its large scale, our dataset features a number of in-the-wild challenges—including dynamic objects, lighting variations, motion blur, and imperfect camera poses. Most importantly, WildCity supports a range of downstream tasks: from reconstructing urban digital twins for photorealistic rendering and closed-loop simulation~\cite{tancik2022block,liu2025citygaussian,kerbl2024hierarchical}, to studying spatial memory, localization, perception, and reasoning with VLMs across a large city~\cite{Liu_2025_citywalker,NEURIPS2020_MultiON}. In this paper, we focus on city-scale rendering as a first concrete instantiation of this broader vision.

Beyond dataset curation, we build a strong baseline tailored to the challenges of large-scale, noisy, and unbounded scenes. We reconstruct urban environments at scale and integrate the resulting models into a closed-loop simulator, demonstrating their utility for downstream embodied tasks. Moreover, we conduct a systematic analysis of the key obstacles in real-world city-scale reconstruction: \textcolor{myred}{\textbf{performance scalability}}, 
\textcolor{mygreen}{\textbf{view extrapolation}}, and 
\textcolor{myblue}{\textbf{data uncertainty}}. This reveals fundamental limitations of existing methods and points toward promising directions for building simulation-ready urban digital twins and, more broadly, advancing city-scale spatial intelligence. Our main contributions are threefold:
\begin{itemize}
\item \textbf{A city-scale real-world dataset.} We introduce WildCity, a large-scale benchmark featuring various in-the-wild challenges—dynamic objects, lighting variations, and imperfect observations—making it a realistic platform for reconstruction, rendering, and embodied AI at city scale.

\item \textbf{An urban-tailored reconstruction baseline.} We establish a baseline specifically designed for large, noisy, and unbounded street scenes, and integrate the resulting reconstructions into a closed-loop simulator to support downstream embodied tasks such as end-to-end autonomous driving.

\item \textbf{A systematic analysis of key challenges.} We identify fundamental limitations of existing pipelines, \ie, scalability, extrapolation, and uncertainty, and outline directions toward simulation-ready urban digital twins.
\end{itemize}

%% file: sec/2_related_works.tex
\section{Related Works}
\label{sec:related_works}

\noindent \textbf{Large-Scale 3D Reconstruction.} Scaling neural rendering from objects and indoor scenes to unbounded outdoor environments remains a significant challenge~\cite{tancik2022block, liu2025citygaussian}. Early NeRF-based methods such as Block-NeRF~\cite{tancik2022block} and Mega-NeRF~\cite{turki2022mega} improve scalability by decomposing large scenes into spatial submodels, but their practical utility remains limited by high training and rendering costs. The emergence of 3D Gaussian Splatting (3DGS)~\cite{kerbl3Dgaussians} has substantially improved the efficiency of large-scale reconstruction. VastGaussian~\cite{lin2024vastgaussian}, CityGaussian~\cite{liu2025citygaussian, liu2024citygaussianv2}, UrbanGS~\cite{li2026urbangs}, and FlashGS~\cite{feng2025flashgs} further enhance optimization and rendering efficiency for large scenes, though they primarily target aerial views or scenes below the scale of entire cities~\cite{lin2024vastgaussian,liu2025citygaussian,liu2024citygaussianv2}. In contrast, real-world street-view city-scale reconstruction involves long and narrow trajectories, limited viewpoint overlap, and substantial sensor noise~\cite{yan2024street,hu2024ss3dm}. More recently, feed-forward 3D reconstruction methods such as DUSt3R~\cite{wang2024dust3r} and VGGT~\cite{wang2025vggt}, together with their longer-horizon extensions~\cite{deng2025vggtlong, wang2025continuous,chen2025ttt3r}, have emerged as a promising alternative by predicting pixel-aligned geometry from learned 3D priors. However, they remain limited in multi-kilometer urban settings due to long-horizon pose drift and the restricted fidelity of sparse point-based representations~\cite{wang2024dust3r,wang2025vggt,wang2025continuous,chen2025ttt3r}.

\noindent \textbf{City-Scale Street-View Datasets.} Existing datasets for urban scene reconstruction can be broadly grouped into synthetic city-scale benchmarks, real-world clip-level driving datasets, and real-world route- or block-level reconstruction datasets (see~\cref{tab:dataset_comparison_recon}). Synthetic datasets such as MatrixCity~\cite{li2023matrixcity} and SS3DM~\cite{hu2024ss3dm} provide broad coverage and clean annotations through controllable simulation, but suffer from sim-to-real gaps~\cite{peng2018visda, safaei2025quantifying, yao2025style}. Real-world driving datasets such as PandaSet~\cite{xiao2021pandaset}, Argoverse 2~\cite{wilson2023argoverse}, WayveScenes101~\cite{zurn2024wayvescenes101}, nuScenes~\cite{caesar2020nuscenes}, and Waymo~\cite{sun2020scalability} contain surround-view observations in real urban environments, but are primarily organized as short clips for perception tasks and therefore lack the continuous city-scale coverage. Route-level datasets such as KITTI-360~\cite{liao2022kitti} and Oxford RobotCar~\cite{maddern2020real} offer longer traversals, yet remain limited in geographic diversity, sensor coverage, or benchmark design for neural rendering. Reconstruction-oriented datasets such as Block-NeRF~\cite{tancik2022block}, Oxford Spires~\cite{tao2024oxford}, and H-3DGS~\cite{kerbl2024hierarchical} move closer to large-scale scene reconstruction, but their geographic diversity and spatial coverage remain limited. \textit{Hence, a benchmark that jointly provides real-world data, surround-view sensing, continuous coverage, and street-view imagery across multiple cities is still missing.}

\begin{table*}[t]
\centering
\scriptsize
\setlength{\tabcolsep}{3.2pt}
\renewcommand{\arraystretch}{0.4}
\begin{tabular}{lccccccc}
\toprule
\textbf{Dataset} 
& \textbf{Scale}
& \textbf{Real}
& \textbf{\#Cities}
& \makecell{\textbf{Coverage} \\ \textbf{(Avg / Total km)}}
& \makecell{\textbf{Surround-} \\ \textbf{view}}
& \makecell{\textbf{Continuous} \\ \textbf{Coverage}} \\
\midrule
MatrixCity~\cite{li2023matrixcity}     & City  & \xmark & 2 & -- / 75.3   & \cmark & \cmark \\
SS3DM~\cite{hu2024ss3dm}                     & Route & \xmark & 8 & 0.48 / 13.4 & \cmark & \cmark \\
\midrule
PandaSet~\cite{xiao2021pandaset}               & Clip  & \cmark & 1 & 0.06 / 6.5  & \cmark & \xmark \\
Argoverse 2~\cite{wilson2023argoverse}          & Clip  & \cmark & 6 & 0.12 / 120  & \cmark & \xmark \\
WayveScenes101~\cite{zurn2024wayvescenes101}   & Clip  & \cmark & 1 & 0.16 / 16.0 & \cmark & \xmark \\
nuScenes~\cite{caesar2020nuscenes}     & Clip  & \cmark & 2 & 0.18 / 100  & \cmark & \xmark \\
Waymo Open~\cite{sun2020scalability}   & Clip  & \cmark & 3 & 0.20 / 100  & \cmark & \xmark \\
\midrule
KITTI-360~\cite{liao2022kitti}         & Route & \cmark & 1 & 6.7 / 74    & \pmark & \cmark \\
Oxford RobotCar~\cite{maddern2020real}        & Route & \cmark & 1 & 10.0 / 1010 & \pmark & \cmark \\
\midrule
Block-NeRF~\cite{tancik2022block}      & Block & \cmark & 1 & 1.08 / 1.08 & \cmark & \cmark \\
Oxford Spires~\cite{tao2024oxford}      & Block & \cmark & 1 & 1.25 / 30.0 & \cmark & \cmark \\
H-3DGS~\cite{kerbl2024hierarchical}    & Block & \cmark & 1 & 3.02 / 9.05 & \cmark & \cmark \\
\specialrule{0.08em}{0.15em}{0.08em}
\rowcolor{black!6}
\textbf{WildCity (Ours)}       & \textbf{City} & \textbf{\cmark} & \textbf{6} & \textbf{83.7 / 1507.1} & \textbf{\cmark} & \textbf{\cmark} \\
\bottomrule
\end{tabular}
\caption{\textbf{Comparison with existing street-view urban reconstruction datasets.} WildCity is the only real-world dataset that jointly provides multi-city coverage, long continuous traversals, surround-view sensing, and city-scale route length. ``$\sim$'' indicates partial or limited support.}

\label{tab:dataset_comparison_recon}
\vspace{-8mm}
\end{table*}

\noindent \textbf{Embodied Spatial Reasoning and Navigation.}
Researchers have long studied spatial reasoning and long-horizon decision-making for embodied agents. Early benchmarks such as MultiON~\cite{NEURIPS2020_MultiON} and Habitat-Web~\cite{ramrakhya2022habitat} emphasize memory, object-centric exploration, and large-scale embodied interaction in photorealistic environments, primarily indoors.
More recent works expand toward richer multimodal and longer-horizon tasks. GOAT-Bench~\cite{Khanna_2024_goat} studies lifelong multimodal navigation over sequential open-vocabulary goals, while LH-VLN~\cite{song2025lhvln} targets long-horizon vision-language navigation with multi-stage planning. CityWalker~\cite{Liu_2025_citywalker} further moves toward urban embodied navigation by learning from web-scale city walking and driving videos. In autonomous driving, UniDrive-WM~\cite{xiong2026unidrive} couple visual understanding, and future scene generation for decision-making. However, these works still lack real-world, city-scale digital twins for interactive spatial reasoning and evaluation. In contrast, WildCity provides the sensory foundation for constructing such twins, enabling future research on city-scale spatial memory, localization, reasoning, and agent-based evaluation.

%% file: sec/3_wildcity_dataset.tex
\begin{figure*}[t]
  \centering
  
  \begin{subfigure}{0.33\linewidth}
    \centering
    \includegraphics[width=\linewidth]{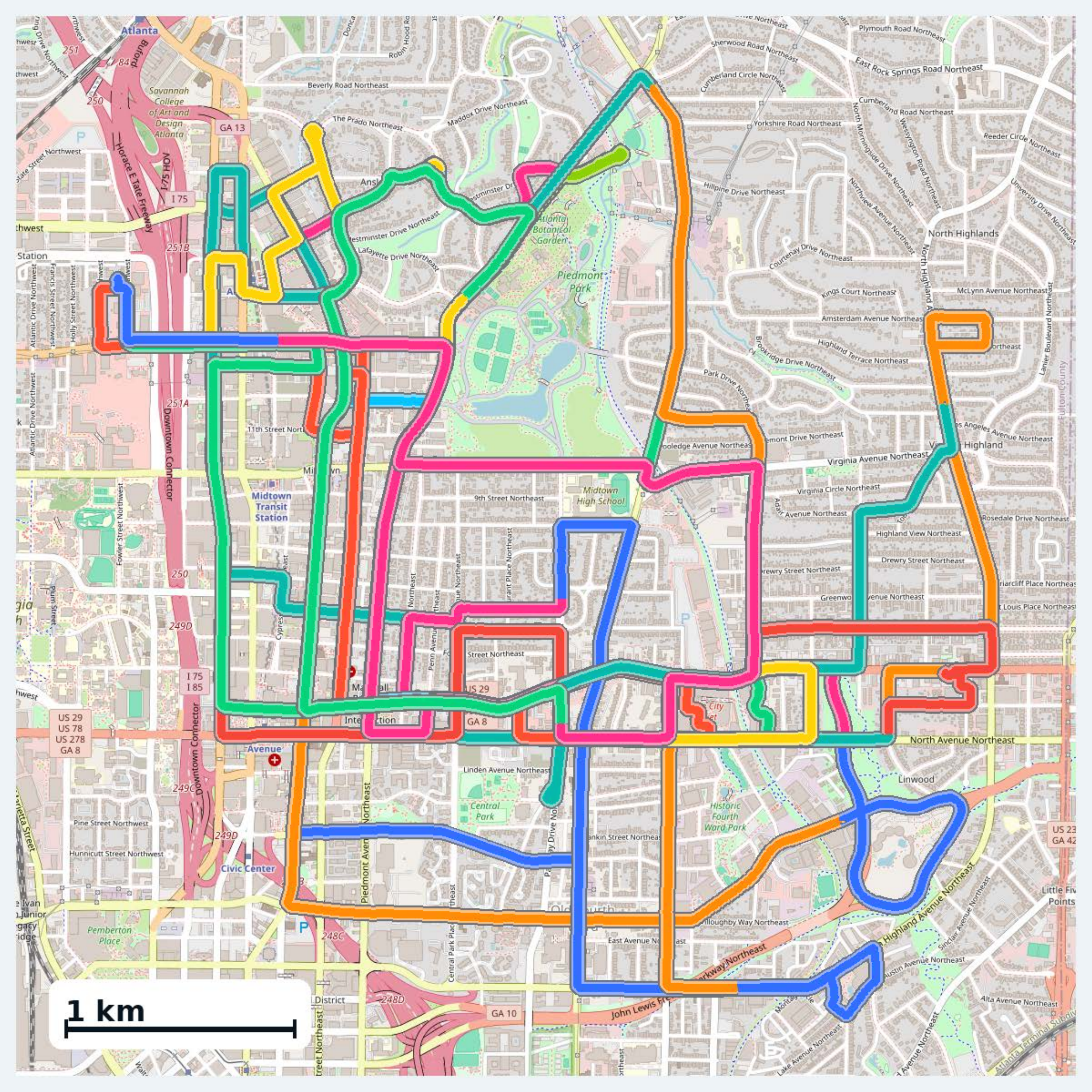}
    \caption{Atlanta}
    \label{fig:sub1}
  \end{subfigure}\hfill
  \begin{subfigure}{0.33\linewidth}
    \centering
    \includegraphics[width=\linewidth]{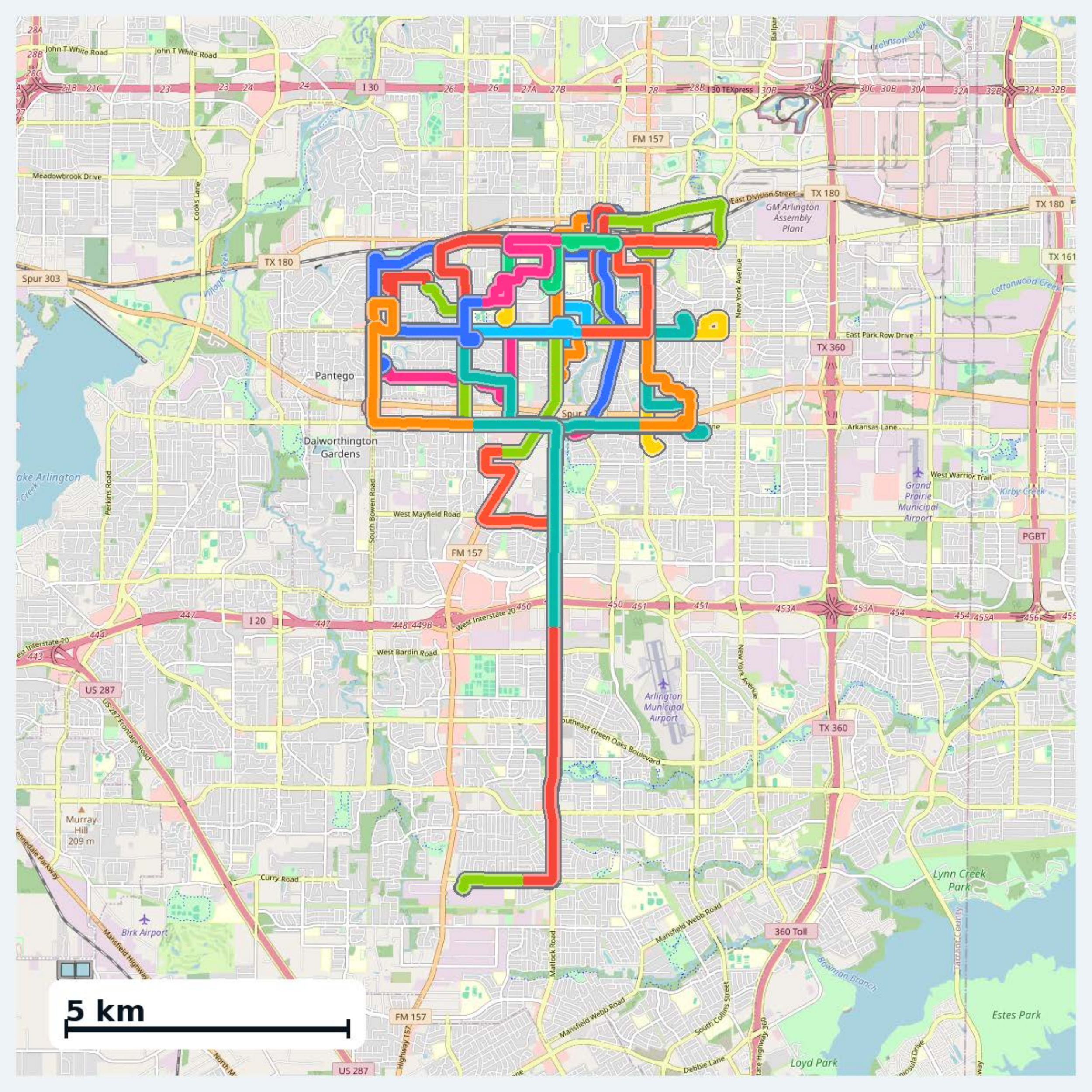}
    \caption{Arlington}
    \label{fig:sub2}
  \end{subfigure}\hfill
  \begin{subfigure}{0.33\linewidth}
    \centering
    \includegraphics[width=\linewidth]{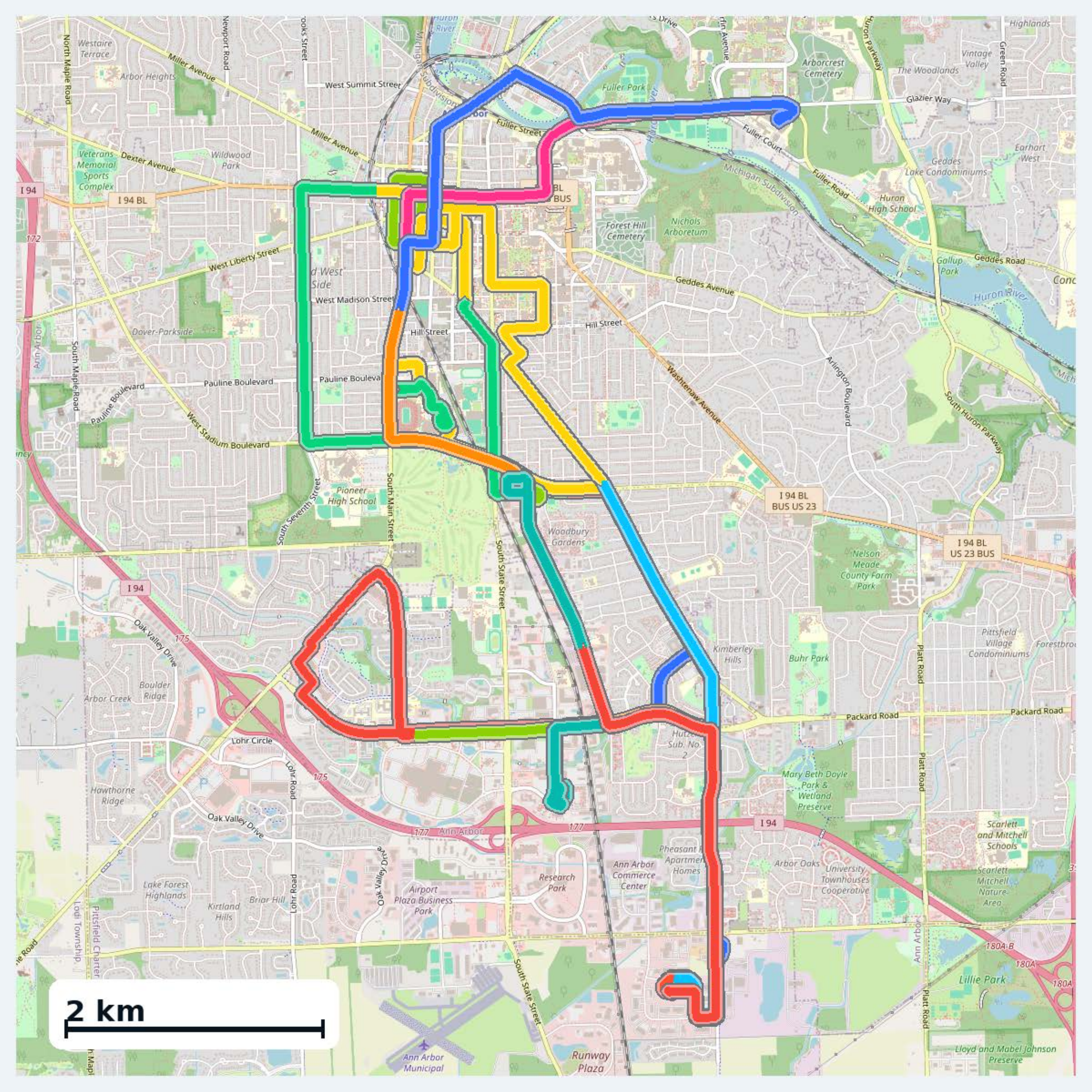}
    \caption{Ann Arbor}
    \label{fig:sub3}
  \end{subfigure}
  
  \vspace{1mm} 
  
  \begin{subfigure}{0.33\linewidth}
    \centering
    \includegraphics[width=\linewidth]{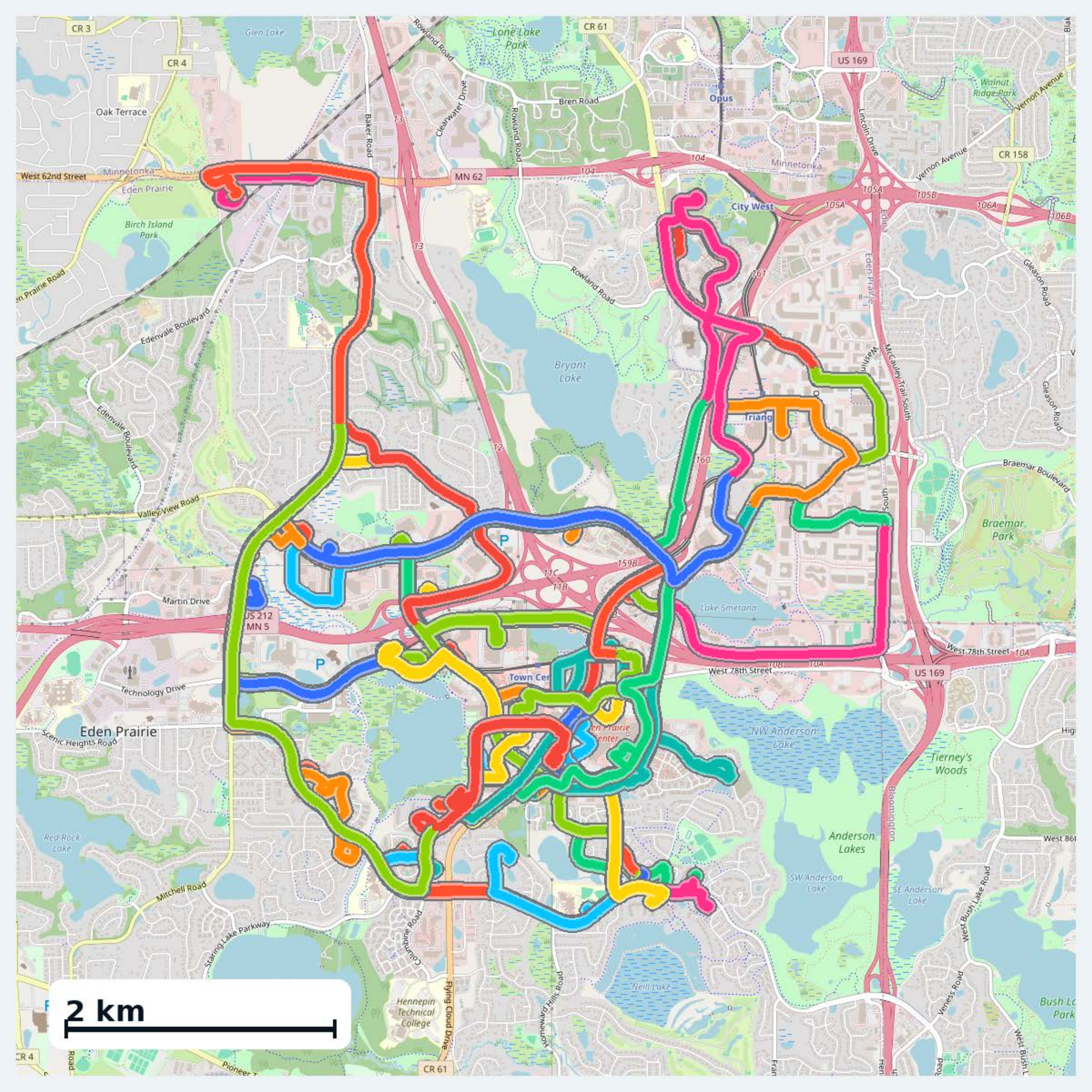}
    \caption{Eden Prairie}
    \label{fig:sub4}
  \end{subfigure}\hfill 
  \begin{subfigure}{0.33\linewidth}
    \centering
    \includegraphics[width=\linewidth]{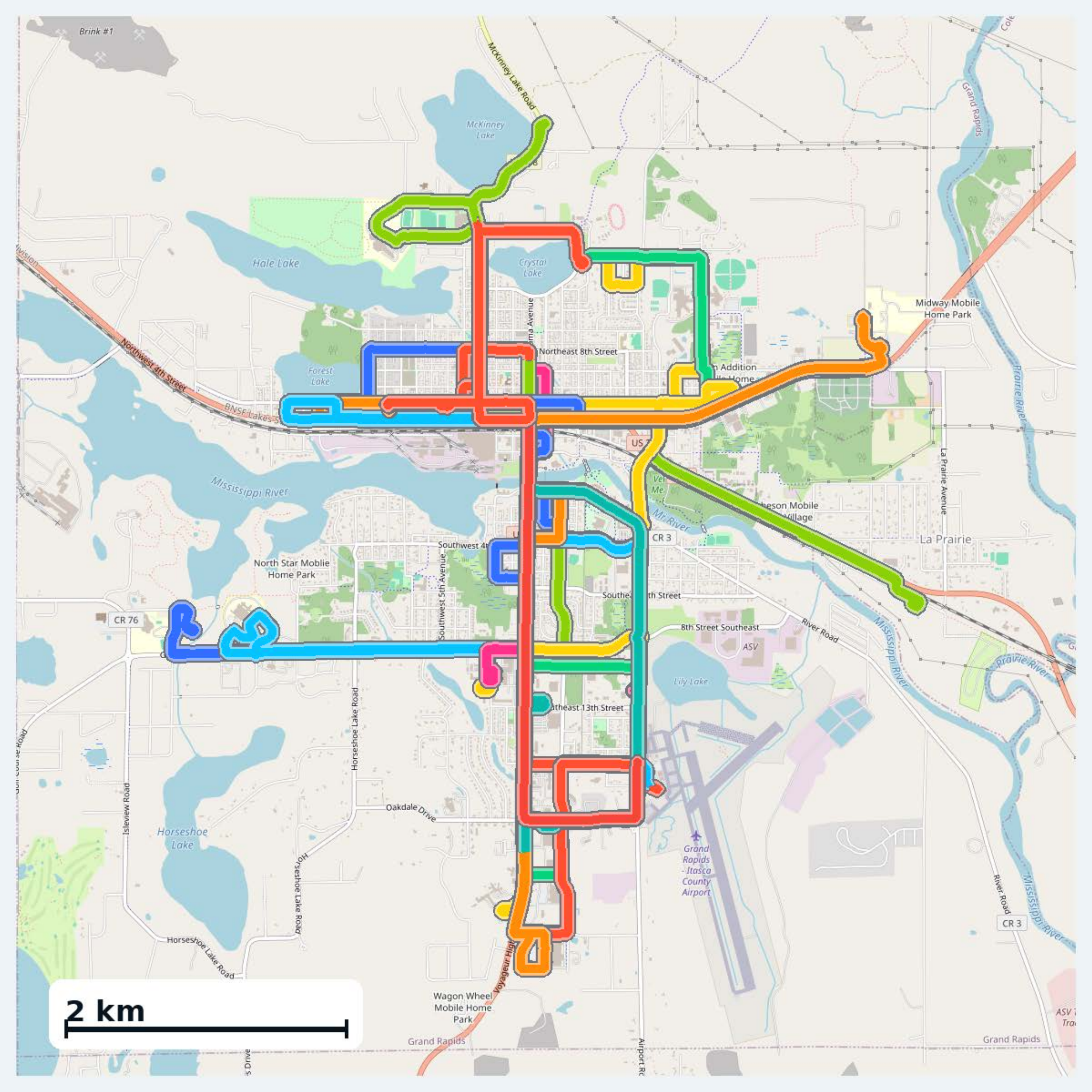}
    \caption{Grand Rapids}
    \label{fig:sub5}
  \end{subfigure}\hfill
  \begin{subfigure}{0.33\linewidth}
    \centering
    \includegraphics[width=\linewidth]{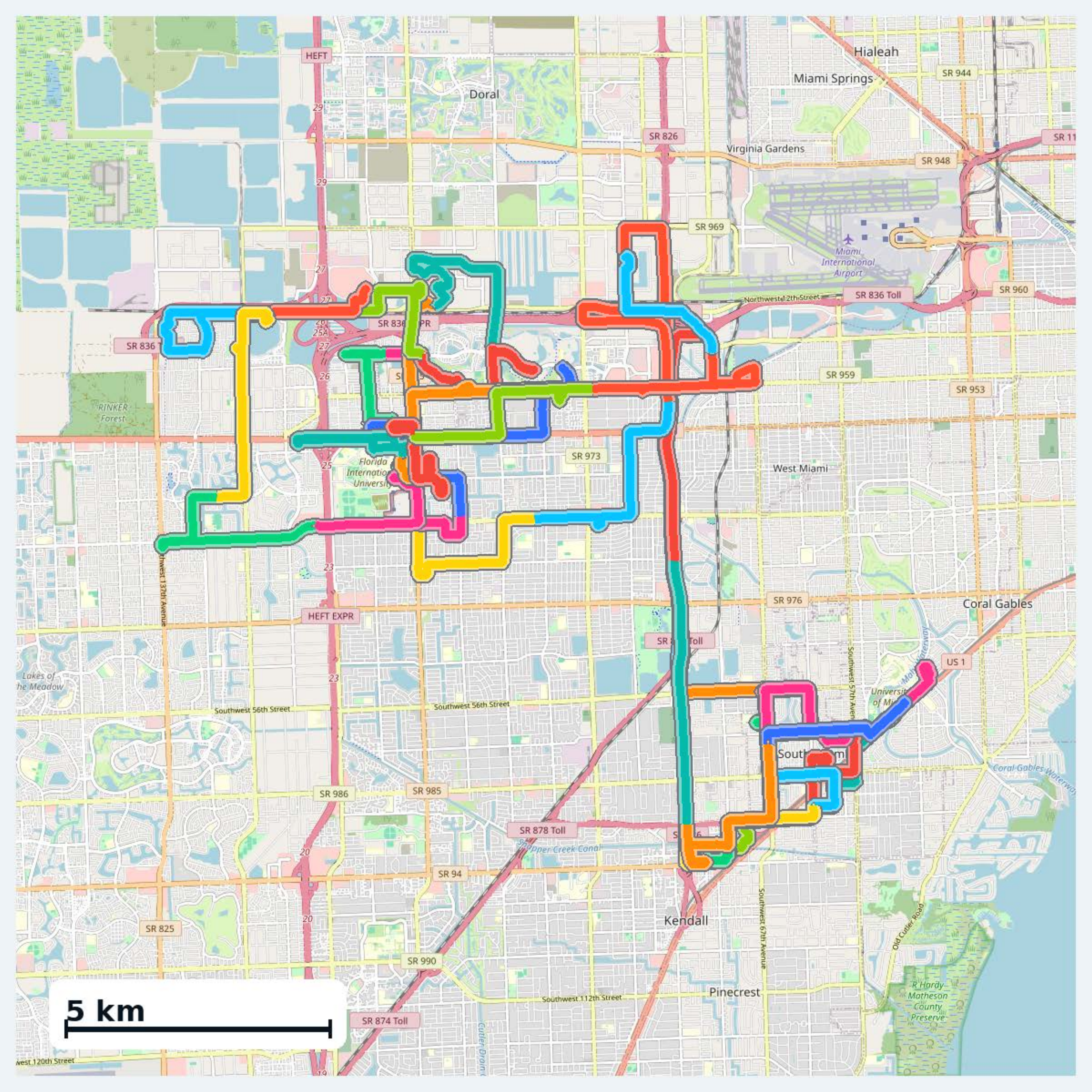}
    \caption{Miami}
    \label{fig:sub6}
  \end{subfigure}
  
    \caption{\textbf{Geographic coverage of \textit{WildCity}.} We visualize trajectories collected across six U.S. cities, with different colors indicating 5-km segments used for standardized data organization and reconstruction. The routes cover diverse urban layouts, including downtown grids, residential neighborhoods, arterial roads, and suburban corridors.}
  \label{fig:geography}
  \vspace{-3mm}
\end{figure*}

\section{WildCity Dataset}
\subsection{Sensor Setup}
\label{Subsec: 3.1}
Our data acquisition platform is equipped with a multi-modal sensor suite comprising a roof-mounted LiDAR, 6 surround-view RGB cameras, an IMU, and a GPS receiver. To ensure comprehensive spatial coverage, the vision system utilizes three narrow-angle cameras for forward-facing views and three wide-angle cameras for lateral and rear views. We provide rigorous calibration, including extrinsic parameters mapping each sensor to the ego-vehicle coordinate frame. Camera intrinsics and lens distortion coefficients are calibrated via AprilCal~\cite{AprilCal}.

\subsection{Data Collection} 
\label{Subsec: 3.2}
Our fleet currently operates in \textit{six U.S. cities: Atlanta, Arlington, Ann Arbor, Eden Prairie, Grand Rapids and Miami}, as shown in~\autoref{fig:geography}. These cities exhibit substantial diversity in urban style, geographic layout, and climate conditions. Across the dataset, the fleet traverses a broad range of street-level scenarios, including dense intersections, narrow local streets, multi-lane arterial roads, and high-speed parkways, under varying traffic patterns and scene complexity. The logs are collected on different days during regular daytime fleet operations, naturally introducing real-world variation in illumination, weather, appearance, and traffic dynamics. As a result, the dataset captures the uncertainty and noise inherent in real urban data, including dynamic objects, appearance changes, and imperfect pose estimates, constructing a challenging but valuable benchmark.

\begin{figure}[t]
    \centering
    \includegraphics[width=0.99\textwidth]{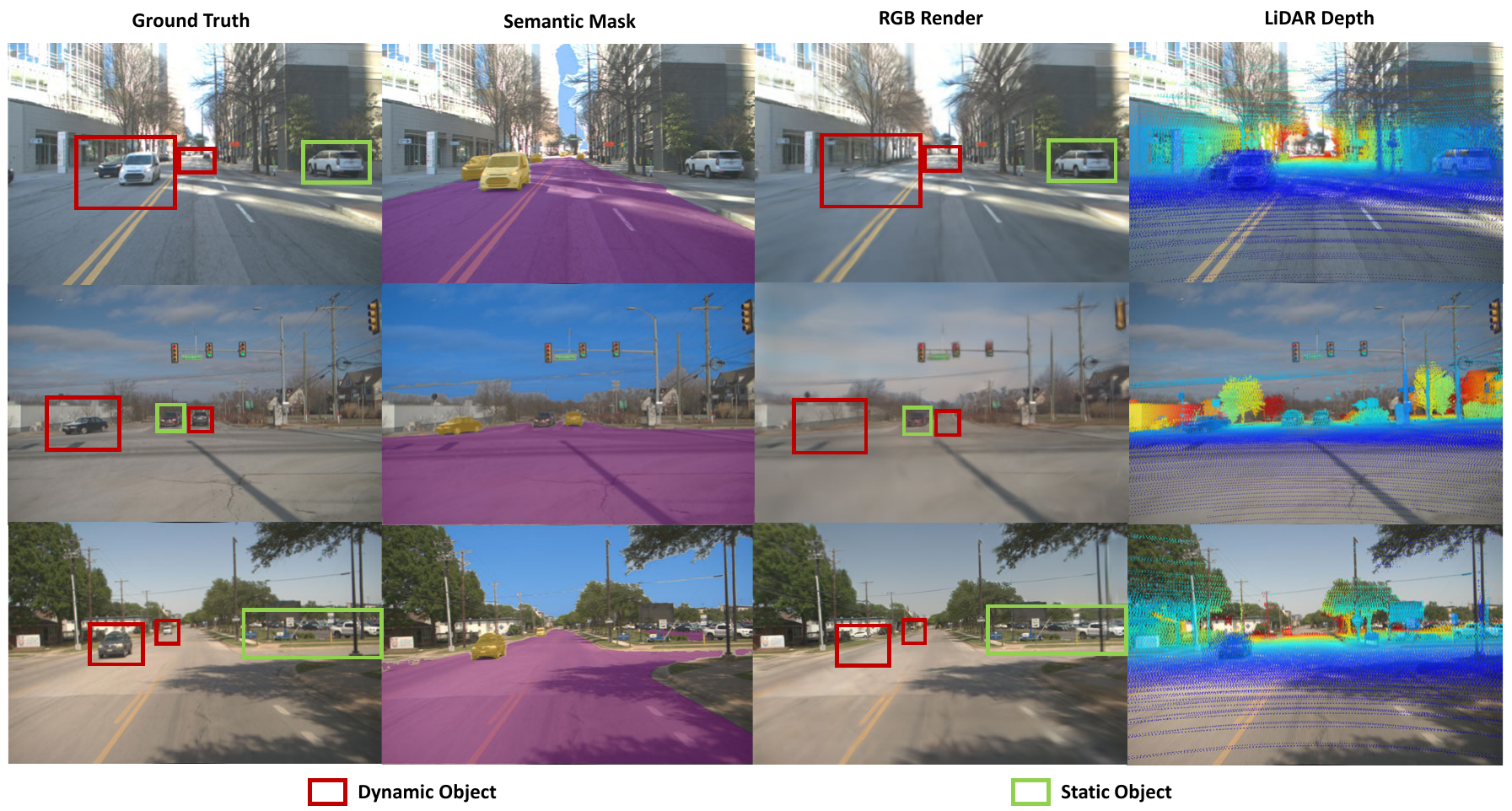}
    \setlength{\fboxsep}{0pt}
    \caption{\textbf{Semantic masks for region-aware reconstruction.} We generate ground, sky, and dynamic-object masks to support road regularization, sky modeling, and moving-object filtering, respectively. For movable categories, 3D tracking cuboids distinguish truly dynamic objects from stationary instances: red boxes indicate objects masked out, while green boxes indicate stationary objects preserved for reconstruction.}

    \label{fig:masks_visualization}
    \vspace{-1em}
\end{figure}

\subsection{Data Processing}
\label{Subsec: 3.3}
The ego poses are initialized from the onboard SLAM system and further refined using GPS signals to improve global consistency, reduce long-range drift, and enforce loop closure when revisiting previously traversed areas. As the raw sensors operate at different frequencies, we extract their measurements directly from the raw logs and align them using the original timestamps instead of resampling them to a fixed frequency. To further improve accuracy, we apply motion compensation to each LiDAR sweep to obtain more accurately aligned geometry. In addition, we provide semantic masks to support broader downstream usage. Specifically, we generate masks for ground, sky, and potentially movable objects (e.g., pedestrians, vehicles, and bicycles) using SAM3~\cite{carion2025sam} with text prompts. We then leverage 3D tracking cuboids from our onboard detection annotations to filter out stationary instances of typically dynamic categories, such as parked vehicles and standing pedestrians. As illustrated in \autoref{fig:masks_visualization}, this process produces region masks that support ground modeling, sky separation, and moving-object filtering for reconstruction. We further validate the generated masks quantitatively and qualitatively in \autoref{tab:mask_miou} and \autoref{fig:mask_examples}, showing that they provide reliable region-level annotations for reconstruction and other tasks that benefit from coarse semantic regions.

\begin{table}[t]
    \centering
    \setlength{\tabcolsep}{10pt}
    \renewcommand{\arraystretch}{0.88}
    \caption{\textbf{Quantitative validation of semantic masks.} We evaluate the automatically generated dynamic-object, sky, and ground masks against 100 manually annotated images across four cities. The high overall mIoU indicates that our mask generation pipeline provides reliable region-level supervision for downstream reconstruction.}
    \vspace{-0.3cm}
    \label{tab:mask_miou}
        \begin{tabular}{lccccc}
        \toprule
        City & \# of Images & Dynamic & Sky & Ground & mIoU \\
        \midrule
        Ann Arbor  & 20 & 83.60 & 96.50 & 95.31 & 91.80 \\
        Atlanta    & 30 & 88.57 & 84.27 & 93.86 & 88.90 \\
        Arlington  & 25 & 88.95 & 93.94 & 94.14 & 92.34 \\
        Eden Prairie      & 25 & 90.11 & 87.62 & 93.95 & 90.56 \\
        \midrule
        Overall    & 100 & 88.48 & 92.01 & 94.24 & 91.58 \\
        \bottomrule
        \end{tabular}
\vspace{-3mm}
\end{table}

\begin{figure}[t]
    \centering
    \setlength{\tabcolsep}{2pt}
    \begin{tabular}{cccc}
        \includegraphics[width=0.235\linewidth]{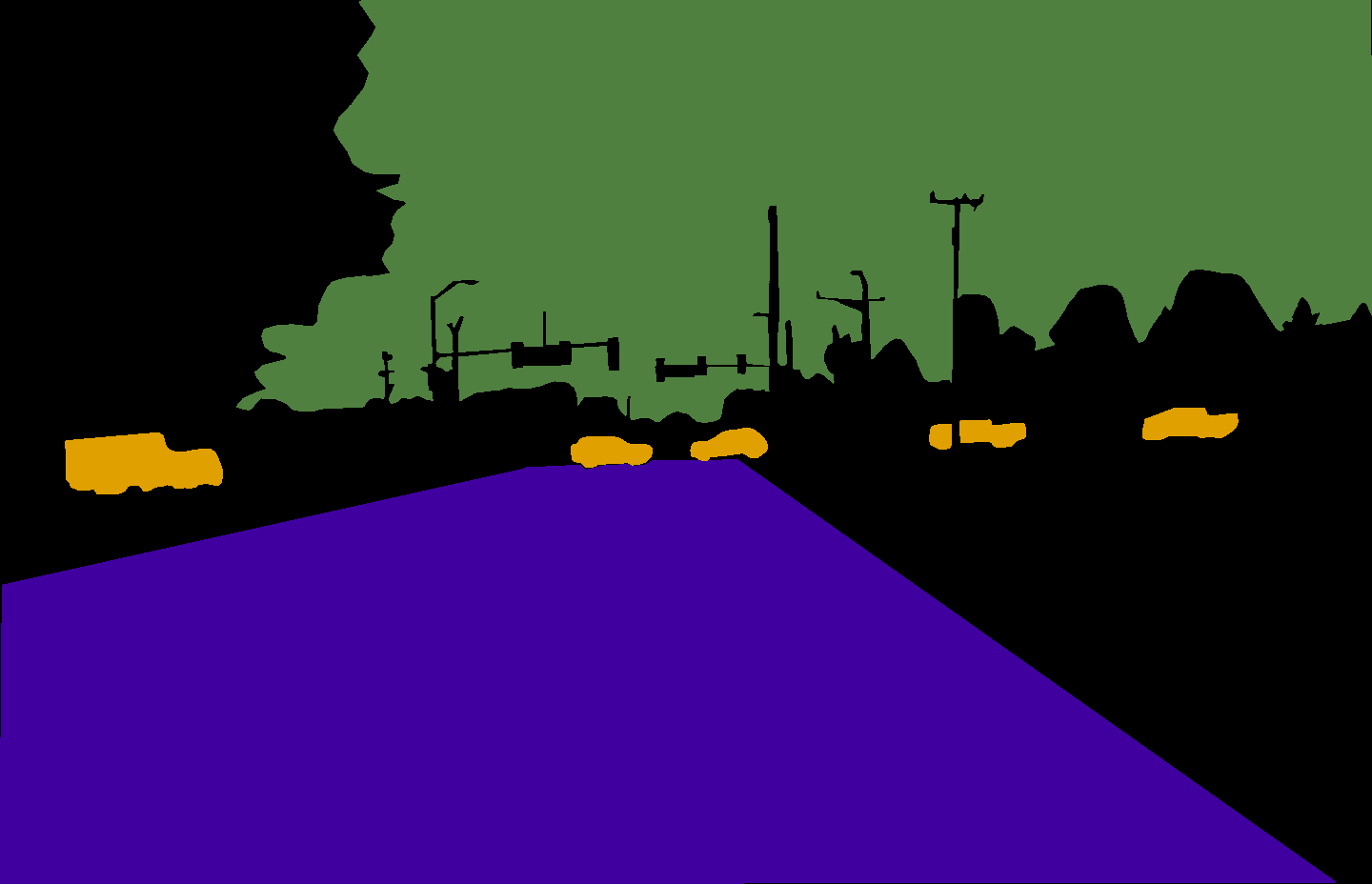} &
        \includegraphics[width=0.235\linewidth]{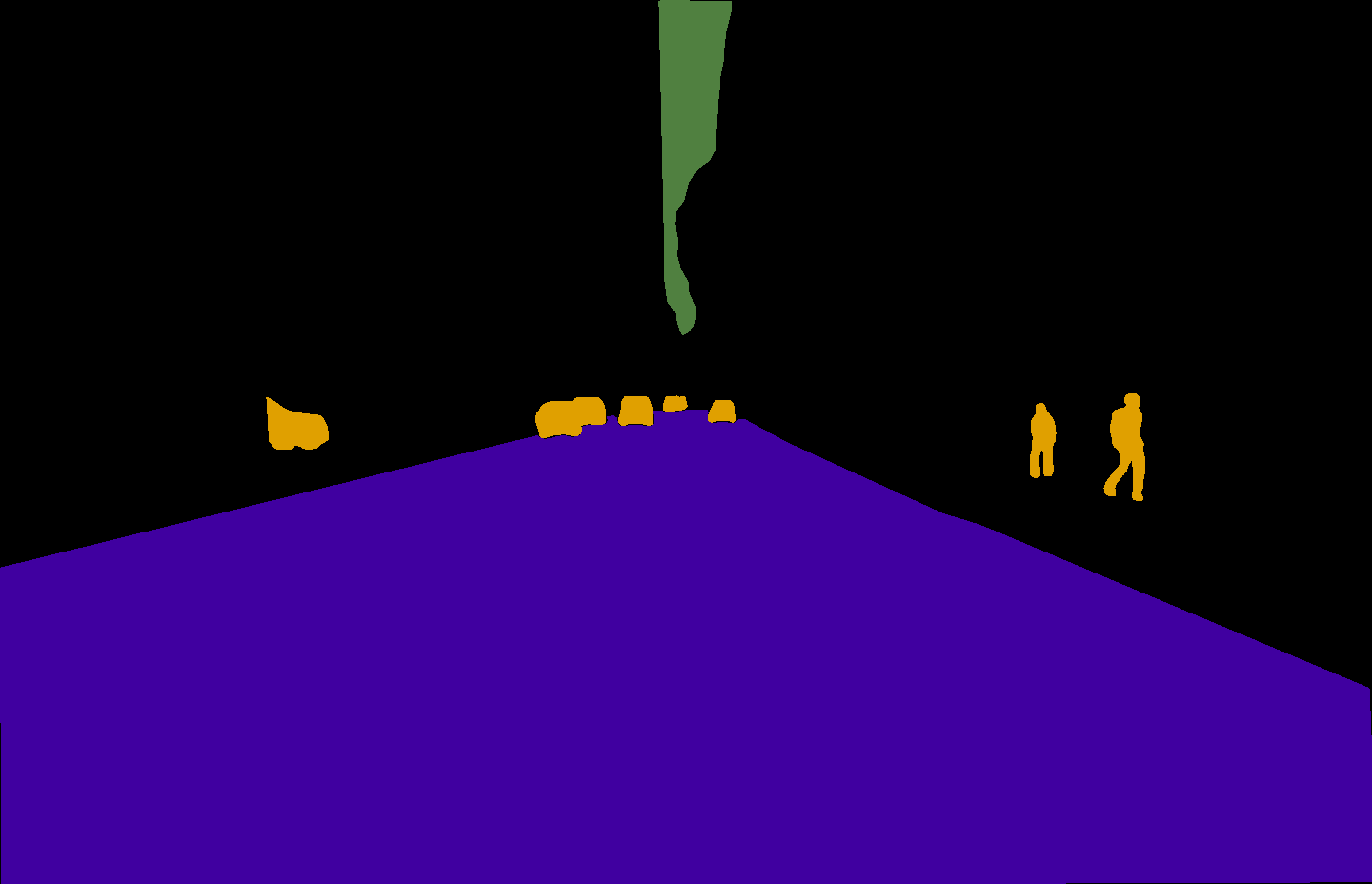} &
        \includegraphics[width=0.235\linewidth]{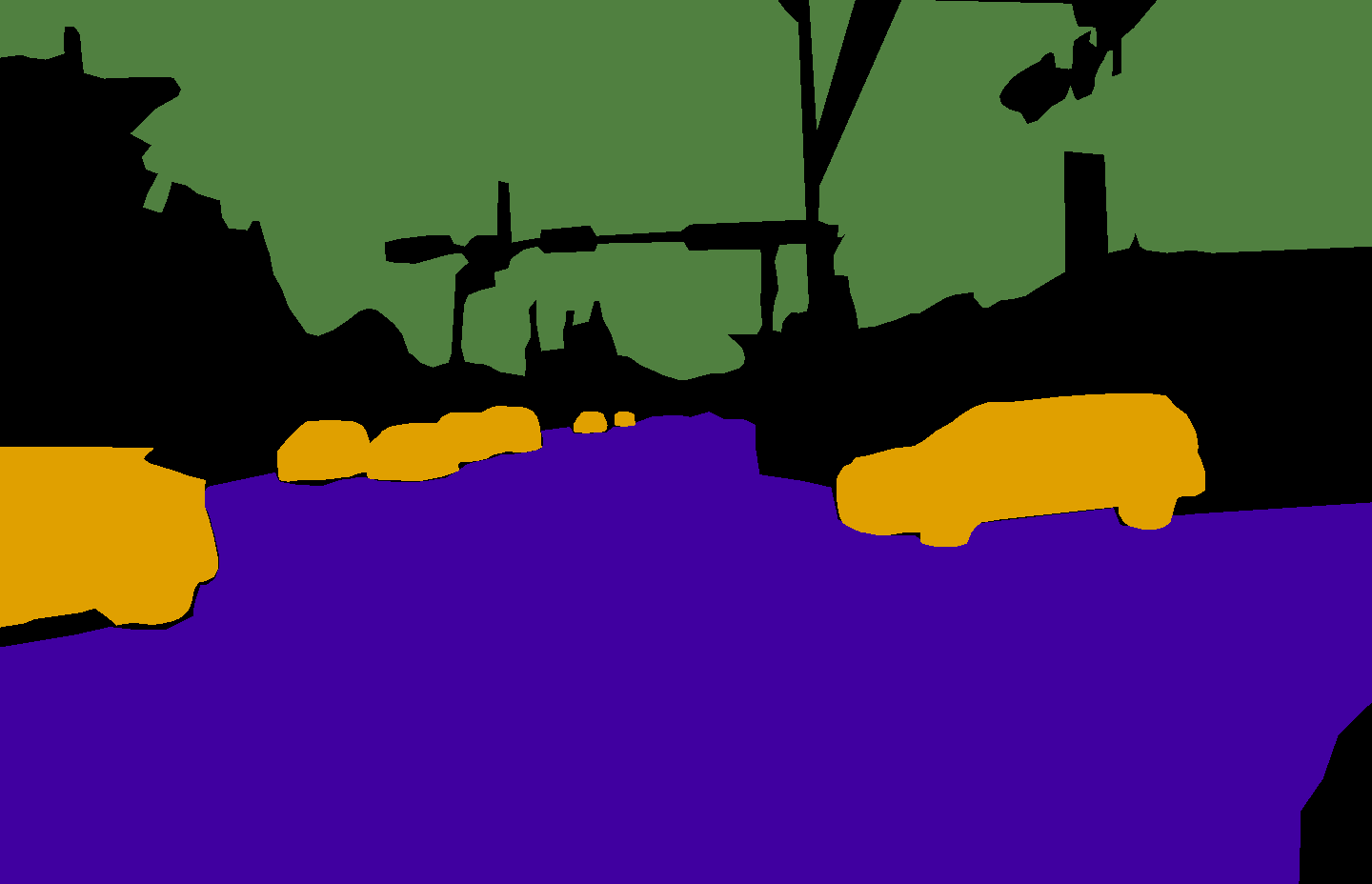} &
        \includegraphics[width=0.235\linewidth]{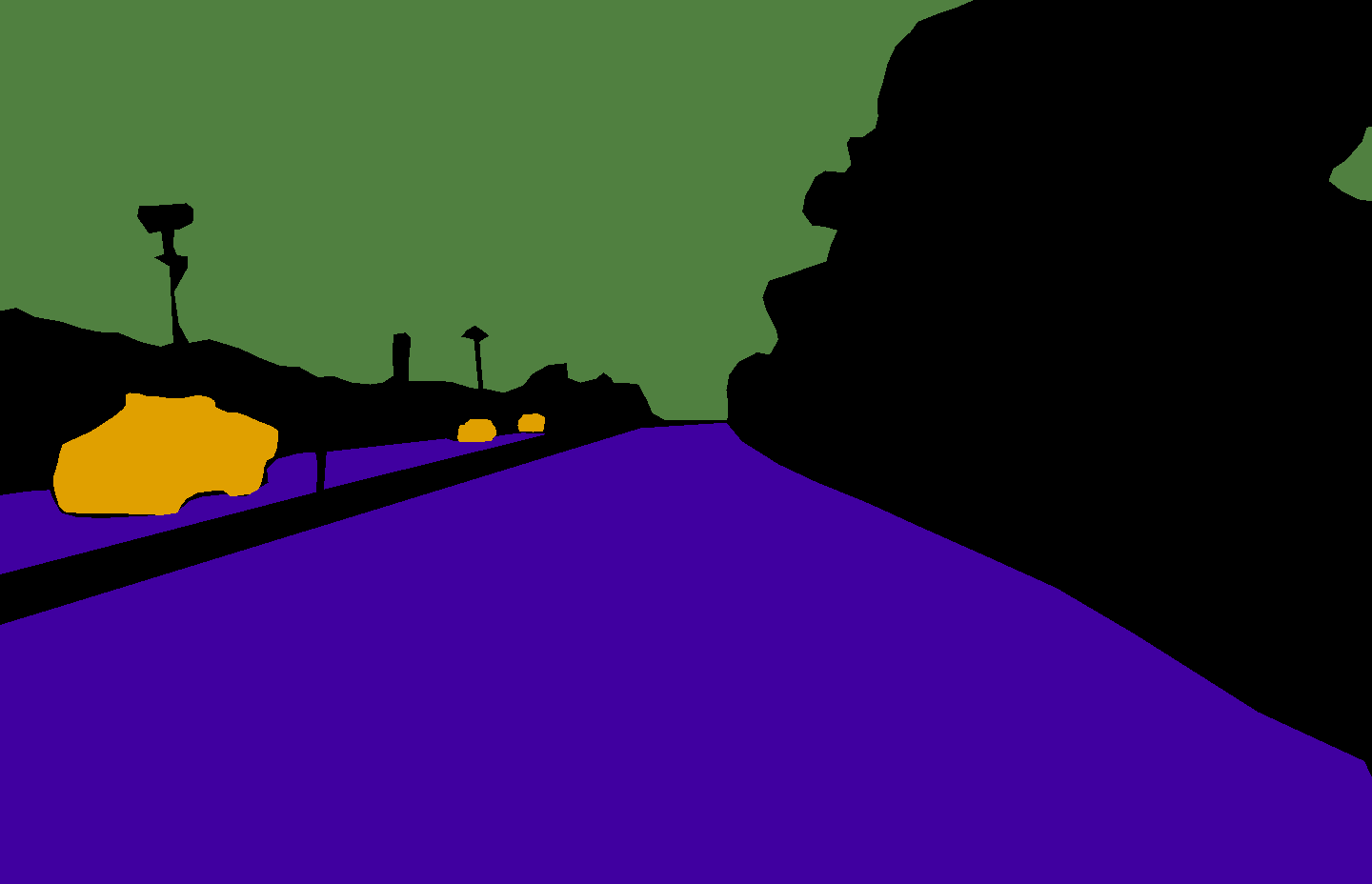} \\
        \includegraphics[width=0.235\linewidth]{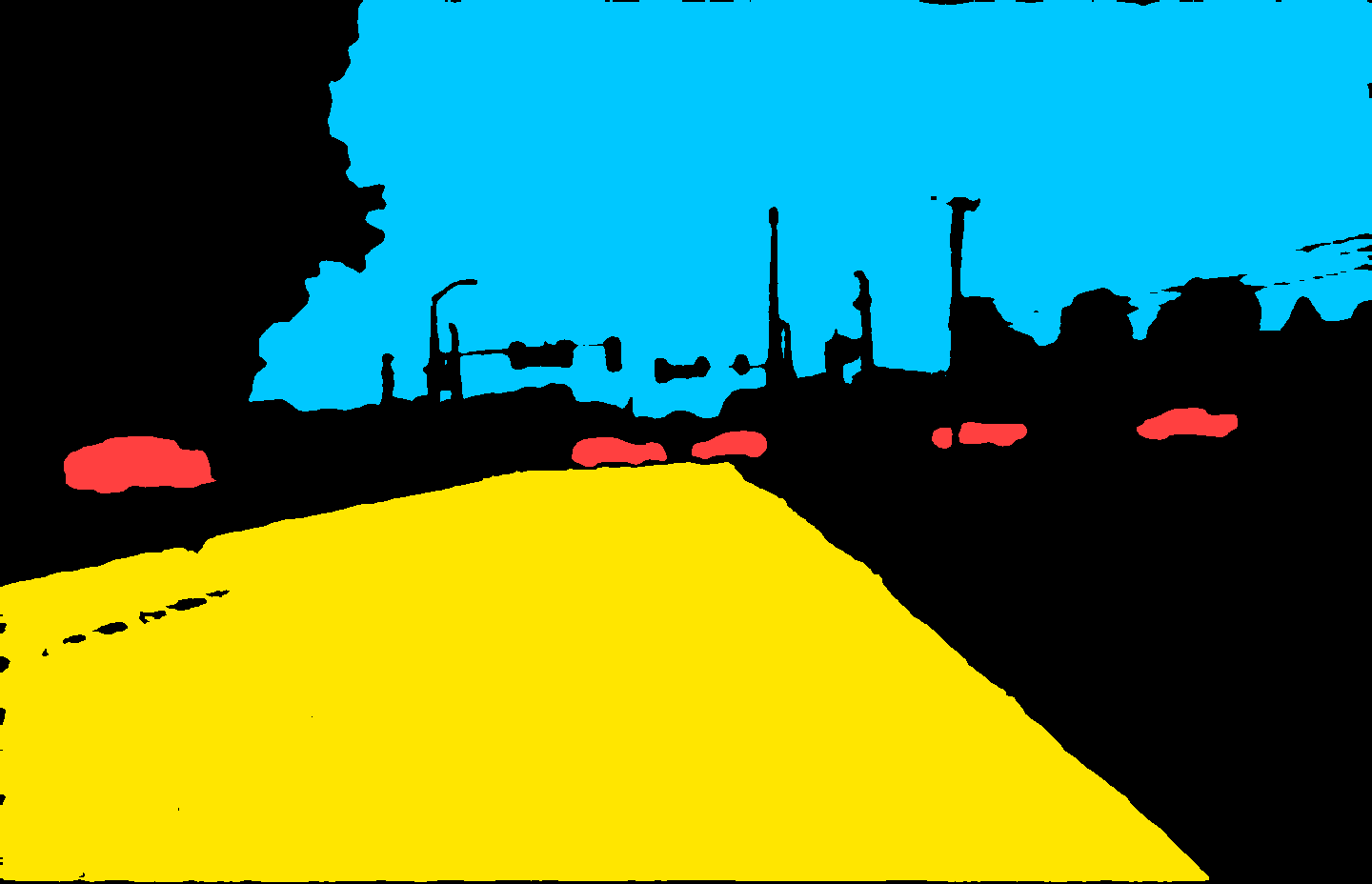} &
        \includegraphics[width=0.235\linewidth]{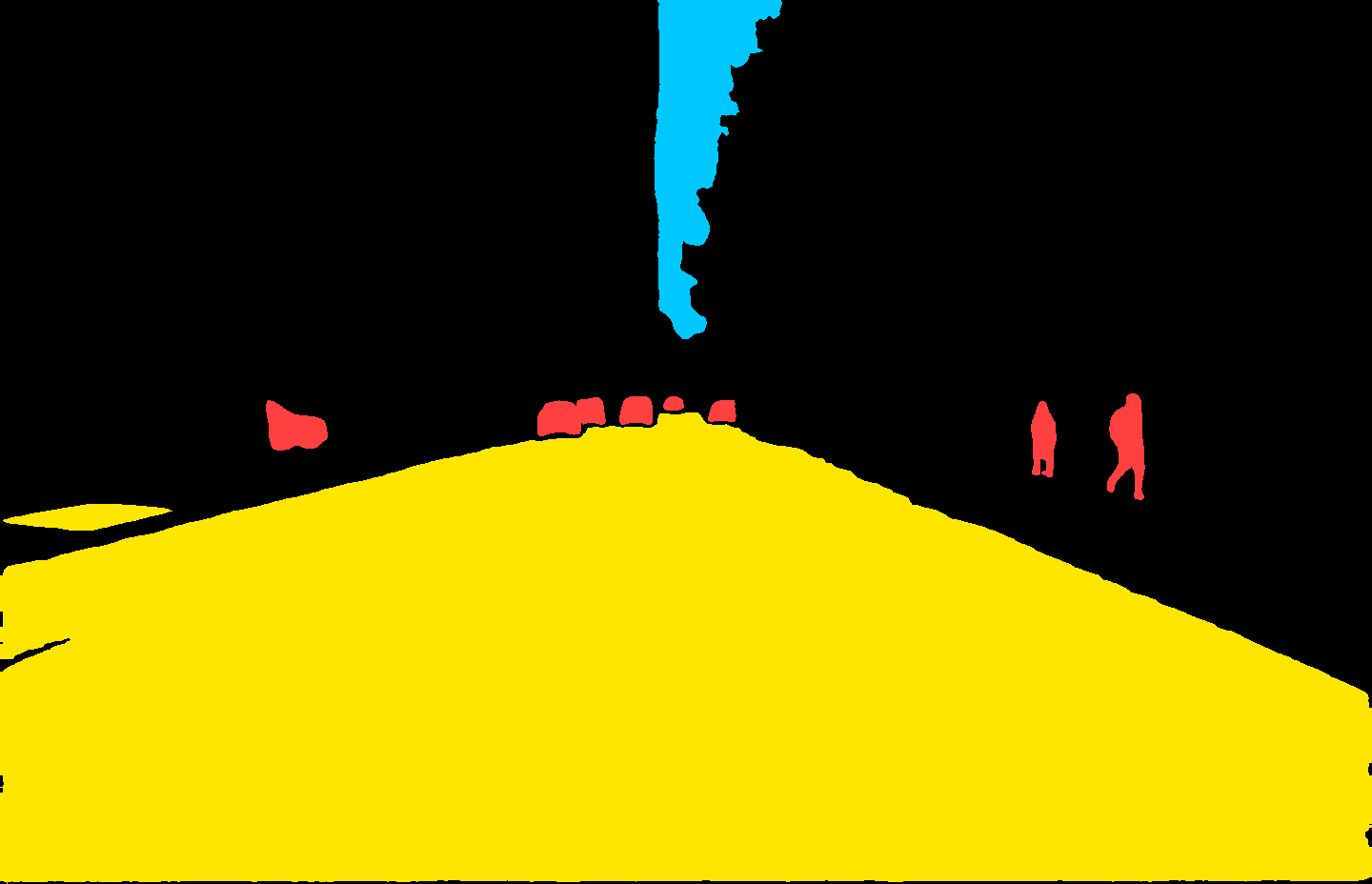} &
        \includegraphics[width=0.235\linewidth]{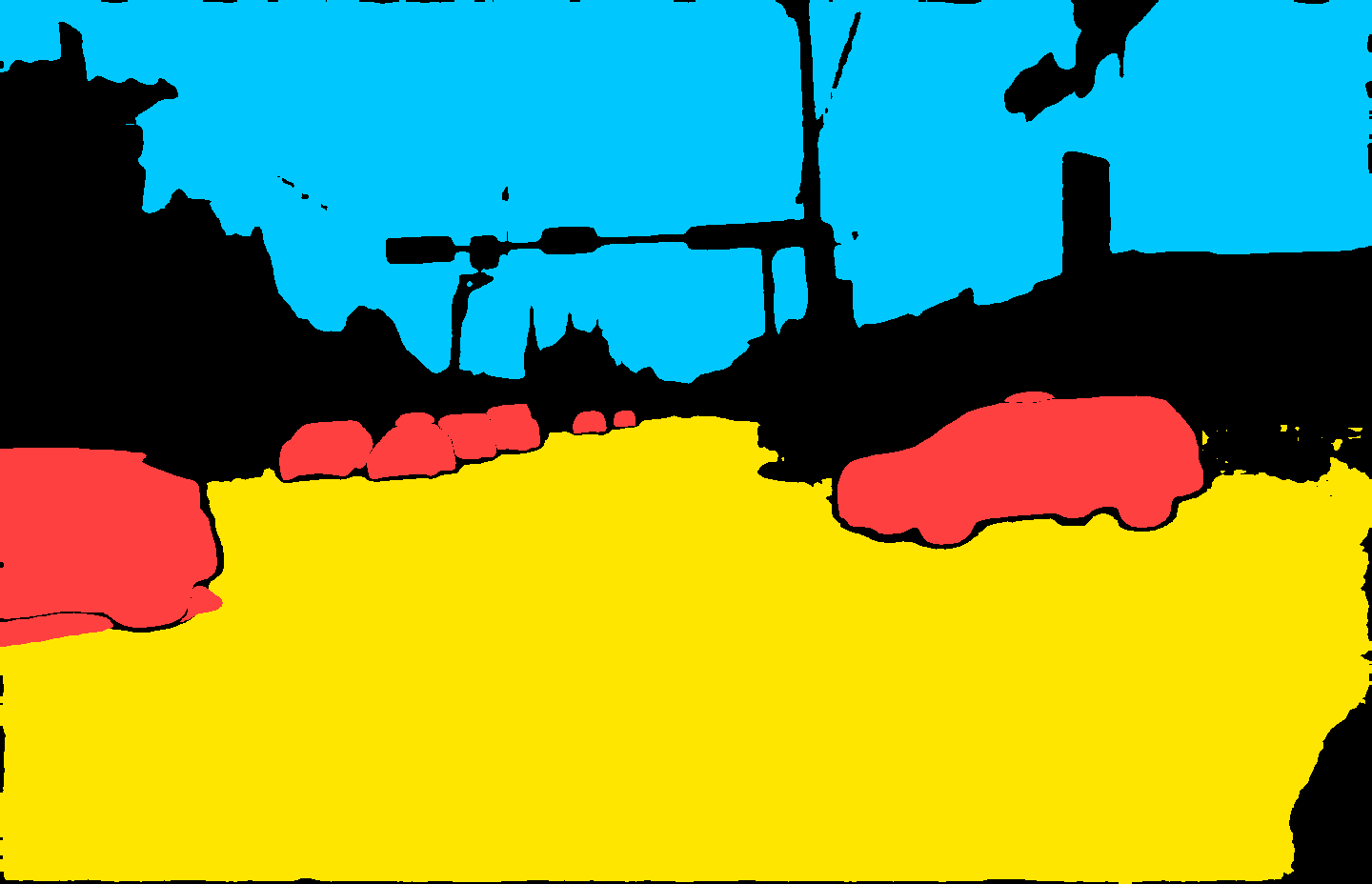} &
        \includegraphics[width=0.235\linewidth]{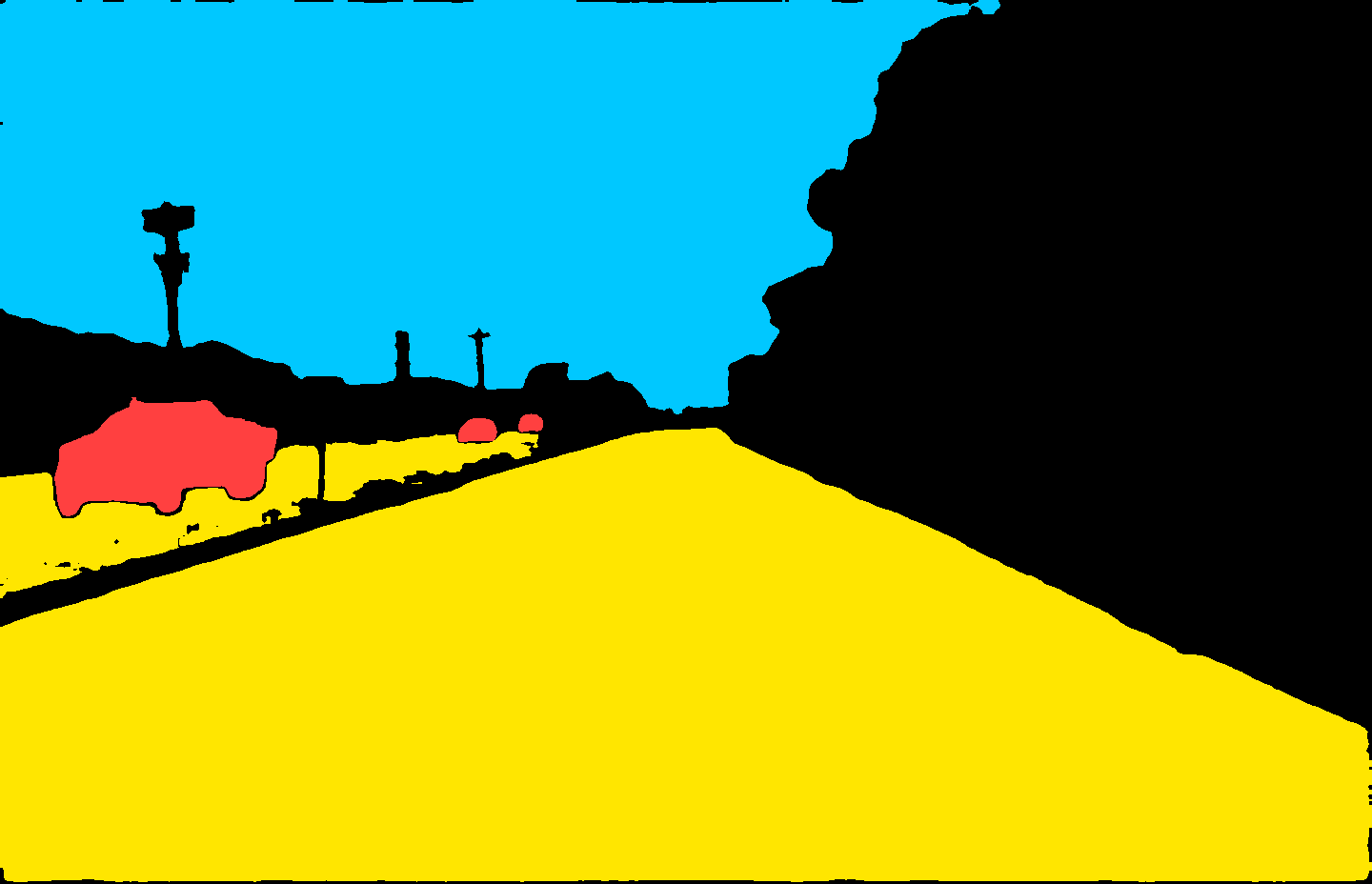}
    \end{tabular}
    \vspace{-0.3cm}
    \caption{\textbf{Qualitative validation of semantic masks.} We compare manual annotations (top) with automatically generated masks (bottom) for dynamic objects, sky, and ground regions. The visual agreement complements the quantitative IoU results in \autoref{tab:mask_miou} and supports the use of automatic masks for region-aware reconstruction.}
    \label{fig:mask_examples}
    \vspace{-0.5cm}
\end{figure}

\subsection{Data Organization}
\label{Subsec: 3.4}
To make long-horizon city logs tractable for reconstruction and benchmarking, we adopt a standardized trajectory discretization and segmentation scheme. We first spatially subsample keyframes at a fixed \textbf{spacing of 0.5\,m} to balance reconstruction fidelity and training efficiency. Each log is then partitioned into contiguous \textbf{5\,km chunks} for convenient usage and reproducible comparisons. On top of these chunks, we provide sub-trajectories at multiple lengths \textbf{[50\,m, 250\,m, 500\,m, 1\,km, 2.5\,km, and 5\,km]} to support evaluation under increasing spatial scale; shorter segments emphasize local consistency while longer segments stress scalability and drift accumulation. Since all poses are aligned in a shared city-level coordinate system, segments can be directly concatenated into longer routes without additional registration.

\begin{table}[t]
\centering

\caption{\textbf{Key dataset and reconstruction statistics.} We summarize the sensor configuration, total data scale, and computational footprint required for reconstruction at increasing trajectory lengths.}
\vspace{-2mm}
\label{tab:dataset_stats}
\scriptsize
\begin{tabular*}{\columnwidth}{@{\extracolsep{\fill}}lc}
    \toprule
    \textbf{Statistic} & \textbf{Value} \\
    \midrule
    Total keyframes & 3.01M \\
    RGB cameras / resolution & 3 @ 1440$\times$928 and 3 @ 1240$\times$728 \\
    LiDAR points per frame & 80k original / 15k downsampled \\
    Camera / LiDAR / GPS / IMU rate & 10 / 10 / 10 / 100 Hz \\
    Avg. GS @ 0.5km / 1km / 2.5km / 5km & 6M / 12M / 30M / 60M \\
    Min. VRAM @ 0.5km / 1km / 2.5km / 5km & 1$\times$24G / 1$\times$40G / 1$\times$80G / 2$\times$80G \\
    \bottomrule
\end{tabular*}
\vspace{-4mm}
\end{table}

\begin{figure}[t]
    \centering
    \includegraphics[width=0.99\textwidth]{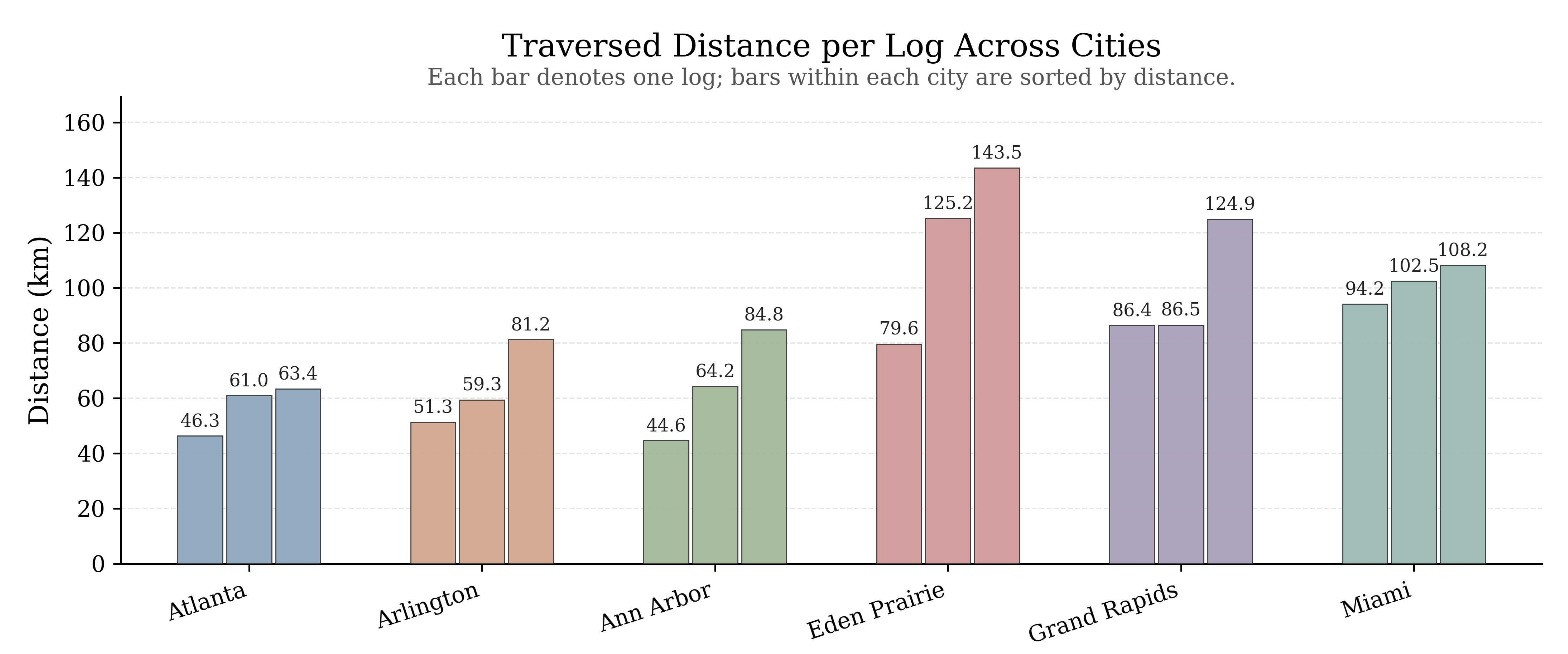}
    \setlength{\fboxsep}{0pt}
    \caption{\textbf{Log-level route coverage.} Each bar represents one continuous driving log, with logs within each city sorted by traversed distance. The distribution highlights the long-horizon, multi-city nature of \textit{WildCity}, in contrast to isolated short-clip datasets.}
    \label{fig:route_statistics}
    \vspace{-1.5em}
\end{figure}

\subsection{Data Statistics}
\label{Subsec: 3.5}
As summarized in \autoref{tab:dataset_stats}, \textit{WildCity} provides \textbf{3.01M keyframes} with synchronized surround-view RGB, LiDAR, GPS, and IMU measurements, along with sensor rates, point-cloud density, and reconstruction resource references to facilitate reproducible training and evaluation. At the route level, the spatial coverage is about \textbf{40.18\,km$^2$} per city, spanning a mixture of downtown regions, residential streets, arterial roads, and other urban corridors. Each driving log lasts about \textbf{2.5 hours} and covers \textbf{83.7\,km} of route length, capturing long continuous traversals rather than isolated short clips. In total, we collect three logs per city, resulting in \textbf{18 logs} and \textbf{1507.1\,km} of recorded driving distance across all cities. As illustrated in \autoref{fig:route_statistics}, the route lengths vary across cities and logs, reflecting the natural diversity of real-world fleet operations rather than an artificially balanced collection process. This organization is particularly important for evaluating methods under realistic city-scale coverage, where both long-range continuity and cross-city diversity matter.

%% file: sec/4_wildcity_baseline.tex
\section{WildCity Method}

\subsection{3D Gaussian Splatting Preliminary}
\label{Subsec: 4.1}
3D Gaussian Splatting (3DGS)~\cite{kerbl3Dgaussians} represents a scene as a set of anisotropic Gaussians \(\mathcal{G}=\{g_i\}_{i=1}^N\), where each Gaussian is parameterized by its opacity, mean position, rotation, scale, and view-dependent color. For rendering, the Gaussians are projected to the image plane and alpha-blended in depth order,
\begin{equation}
C=\sum_{i\in\mathcal{N}} c_i\,\alpha_i \prod_{j=1}^{i-1}(1-\alpha_j),
\end{equation}
where \(\mathcal{N}\) denotes the set of projected Gaussians overlapping the pixel, \(\alpha_i\) is the projected opacity of the \(i\)-th Gaussian and \(c_i\) is the view-dependent color.

\subsection{Urban-tailored Large-scale Reconstruction}
\label{Subsec: 4.2}
\noindent \textbf{Rig pose optimization.} Accurate poses are essential for high-quality neural reconstruction and rendering, but real-world pose estimates from raw logs are often not globally optimal. Although independent camera pose optimization is widely adopted~\cite{chen2024omnire}, it ignores the rigid multi-camera setup at each keyframe and can lead to locally improved but globally inconsistent solutions. We therefore decompose pose optimization into two coupled components: \textit{(1) the ego pose at each keyframe (2) the camera extrinsics relative to the ego frame.} Formally, let $\mathbf{T}^{\mathrm{ego}}_t \in SE(3)$ denote the ego pose at keyframe $t$, and let $\mathbf{T}^{\mathrm{rig}}_c \in SE(3)$ denote the extrinsic pose of camera $c$ in the ego frame. The resulting camera pose is
\begin{equation}
\mathbf{T}_{t,c} = \mathbf{T}^{\mathrm{ego}}_t \mathbf{T}^{\mathrm{rig}}_c.
\end{equation}
We then jointly optimize the scene parameters $\theta$, ego poses, and rig extrinsics via
\begin{equation}
\min_{\theta,\{\mathbf{T}^{\mathrm{ego}}_t\},\{\mathbf{T}^{\mathrm{rig}}_c\}}
\sum_{t,c}
\mathcal{L}_{\mathrm{rend}}\!\left(\mathbf{I}_{t,c},
\mathcal{R}(\theta,\mathbf{T}_{t,c}
\right)
+ \lambda_{pose} (\sum_t d(\mathbf{T}^{\mathrm{ego}}_t,\bar{\mathbf{T}}^{\mathrm{ego}}_t)
+ \sum_c d(\mathbf{T}^{\mathrm{rig}}_c,\bar{\mathbf{T}}^{\mathrm{rig}}_c)),
\end{equation}
where $\mathcal{R}$ denotes the renderer, $\mathbf{I}_{t,c}$ is the image from camera $c$ at keyframe $t$, $\mathbf{\lambda_{pose}}$ represents the pose loss weight, $\bar{\mathbf{T}}^{\mathrm{ego}}_t$ and $\bar{\mathbf{T}}^{\mathrm{rig}}_c$ are the initial poses from localization and calibration, and $d(\cdot,\cdot)$ is a pose-distance metric on $SE(3)$.

\noindent \textbf{Sky model.}
Following a widely used design in neural rendering, we model the sky separately instead of representing it with 3D Gaussians~\cite{chen2024omnire,yan2024street}. Specifically, we use a lightweight view-dependent MLP to predict the sky color image $\mathbf{C}_{\mathrm{sky}}$ and composite it with the rendered Gaussian image $\mathbf{C}_{g}$ as an infinite background:
\begin{equation}
\mathbf{C}=\mathbf{C}_{g}+\left(1-\mathbf{O}_{g}\right)\mathbf{C}_{\mathrm{sky}},
\end{equation}
where $\mathbf{O}_{g}$ denotes the rendered Gaussian opacity. This decouples sky appearance from scene geometry and helps reduce spurious far-depth floaters.

\noindent \textbf{Ground regularization.}
To stabilize underconstrained road geometry, we impose priors on Gaussians identified as ground. We regularize ground height standard deviation by sampling slices along the camera-depth axis and penalizing the variation of the camera-frame vertical coordinate within each slice:
\begin{equation}
\mathcal{L}_{\mathrm{dist}} =
\frac{1}{|\mathcal{K}|}
\sum_{k \in \mathcal{K}}
\mathrm{Std}\!\left(\left\{\mu_{i,y}^{\mathrm{cam}} \mid i \in \mathcal{S}_k \right\}\right).
\end{equation}
where $\mathcal{K}$ is the set of local ground Gaussians, $\mu_{i,y}^{\mathrm{cam}}$ is the position of splats in camera y axis. We further encourage ground Gaussians to be vertically aligned and sufficiently opaque:
\begin{equation}
\qquad
\mathcal{L}_{\mathrm{align}} =
\frac{1}{|\mathcal{G}|}
\sum_{i \in \mathcal{G}}
\left(1 - \left|a_{i,y}^{\mathrm{cam}}\right|\right),
\qquad
\mathcal{L}_{\mathrm{opa}} =
\frac{1}{|\mathcal{G}|}
\sum_{i \in \mathcal{G}}
\left(1 - o_i \right)^2.
\end{equation}
The final loss is
\begin{equation}
\mathcal{L}_{\mathrm{ground}}
=
\lambda_{\mathrm{dist}}\mathcal{L}_{\mathrm{dist}}
+
\lambda_{\mathrm{align}}\mathcal{L}_{\mathrm{align}}
+
\lambda_{\mathrm{opa}}\mathcal{L}_{\mathrm{opa}}
,
\end{equation}
where $\mathcal{G}$ is the set of ground Gaussians, $\mathcal{S}_k$ is the $k$-th sampled depth slice, $a_{i,y}^{\mathrm{cam}}$ the shortest axis component of camera vertical axis and $o_i$ the opacity.

\noindent \textbf{Extrapolated view post-repair.}
Extrapolated rendering often suffers from broken geometry and floating artifacts~\cite{Han_2025_ICCV}. We therefore adopt a progressive render-repair-augment scheme with Difix3D+~\cite{wu2025difix3d+}. Given Gaussian parameters $\theta^{(m)}$ and a sampled extrapolated pose $\tilde{\mathbf{T}}$, we first render
\begin{equation}
\tilde{\mathbf{I}}^{(m)}=\mathcal{R}(\theta^{(m)},\tilde{\mathbf{T}}),
\end{equation}
then repair it with Difix3D+,
\begin{equation}
\hat{\mathbf{I}}^{(m)}=\mathcal{F}_{\text{Difix}}(\tilde{\mathbf{I}}^{(m)}, \mathbf{I}_{\text{ref}}, prompt),
\end{equation}
and add the repaired supervision back to the training set:
\begin{equation}
\mathcal{D}_{m+1}=\mathcal{D}_{m}\cup\{(\tilde{\mathbf{T}},\hat{\mathbf{I}}^{(m)})\}.
\end{equation}
The Gaussians are then re-optimized on the augmented set. Repeating this process progressively repairs geometric artifacts under extrapolated viewpoints.

\noindent \textbf{Multi-GPU training.}
As scene scale increases, the number of Gaussians grows rapidly and training becomes GPU-memory constrained. Following GrendelGS~\cite{zhao2024scaling} and GSplat~\cite{ye2025gsplat}, we enable multi-GPU training by sharding Gaussian parameters across devices and synchronizing the necessary statistics and gradients, allowing us to scale to billion-level primitives with bounded per-GPU memory.

\begin{figure}[t]
    \centering
    \includegraphics[width=0.99\textwidth]{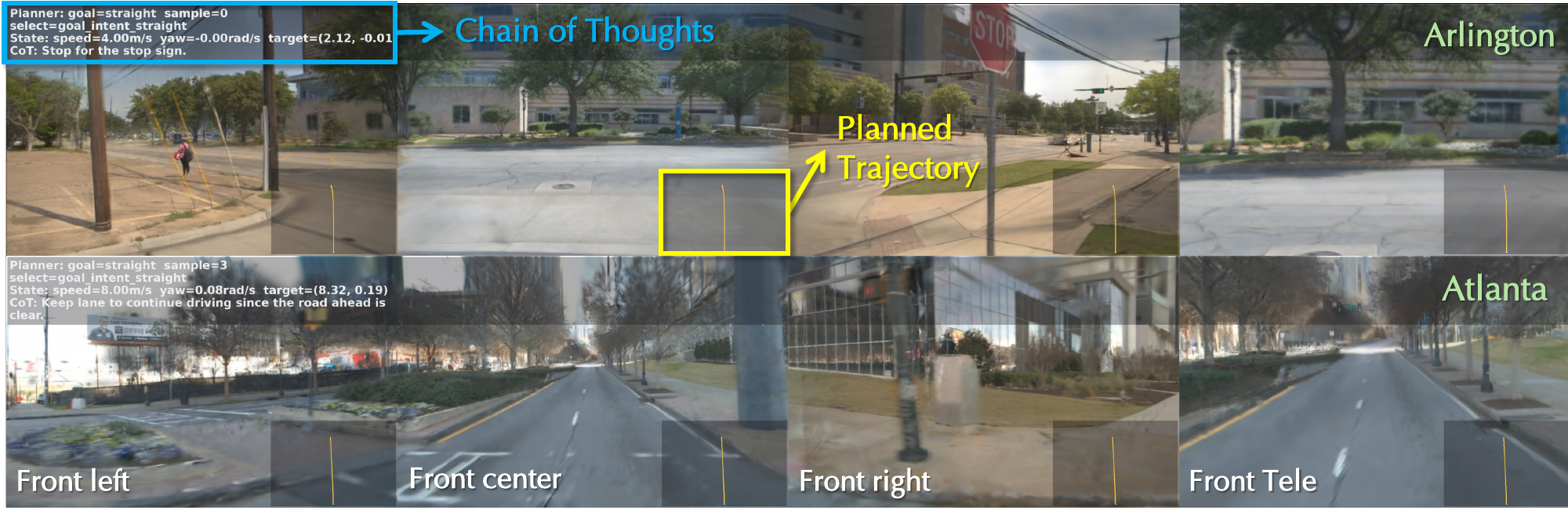}
    \setlength{\fboxsep}{0pt}
    \caption{\textbf{Agents closed-loop simulation.} As proof of concept, Alpamayo~\cite{nvidia2025alpamayo} is integrated into our closed-loop simulator and able to perceive, reason and act interactively. }
    \label{fig:agents_vis}
    \vspace{-1em}
\end{figure}

\subsection{Closed-loop Simulation in City Digital Twin}
\label{Subsec: 4.3}
Beyond reconstruction, \textit{WildCity} enables embodied interaction in a city digital twin, as shown in~\autoref{fig:agents_vis}. A Vision-Language-Action model iteratively predicts the next target pose from the current observation and task, and receives the rendered view at that pose. Using Alpamayo 1~\cite{nvidia2025alpamayo} as a proof of concept, we observe smooth navigation and plausible multi-step reasoning in the reconstructed urban environment. This highlights the broader value of WildCity as a foundation for studying large-scale spatial intelligence in realistic cities.

%% file: sec/5_experiments.tex
\section{Experiments}

\subsection{Evaluation Protocol}
We evaluate 3D reconstruction on \textit{WildCity} across different spatial scales to assess state-of-the-art methods under both local and long-horizon settings. We use two representative sequences captured with the same sensor setup: Ann Arbor-0.5k and Atlanta-5k. We report PSNR, SSIM~\cite{wang2004ssim}, and LPIPS~\cite{zhang2018unreasonable} on the static region defined by semantic segmentation masks, and Depth L1 (meters) on pixels with valid LiDAR depth in the same region. For VGGT-based variants, we align the VGGT reconstruction to our metric pointcloud by optimizing a global similarity transform. Each method is trained under identical and sufficiently provisioned H200 hardware, until validation performance stabilizes when feasible.

\subsection{Comparison with Baselines}
\label{sec:baselines}

\noindent \textbf{Baselines.}
We compare our method against the following baselines: classic 3D Gaussian Splatting (3DGS) \cite{kerbl3Dgaussians}, the scalable hierarchical variant H-3DGS \cite{kerbl2024hierarchical}, and the city-scale blockwise training approach CityGaussianV2 (CityGS) \cite{liu2024citygaussianv2}. We additionally include the feed-forward long-sequence geometry prior VGGT-Long \cite{deng2025vggtlong} as a baseline for point-cloud and pose reconstruction. We further include VGGT-Long+CityGS, where we replace the default SLAM-initialized poses and sparse point clouds in the CityGS training pipeline with the aligned VGGT-Long reconstruction outputs. This baseline tests whether feed-forward priors can serve as a substitute for sensor-based slam initializations on long sequences.

\noindent \textbf{Quantitative comparison.}
As shown in \autoref{tab:baseline_comparison}, while baselines achieve strong 2D image metrics on short sequences, they compromise the underlying 3D geometry. For instance, on the short trajectory, methods like H-3DGS and CityGS report high PSNR or SSIM but exhibit massive geometric distortions, with Depth L1 exceeding \textbf{17\,m}. Our method corrects this misalignment, significantly reducing the depth error to \textbf{11.75\,m} while maintaining highly competitive perceptual quality. This confirms that relying solely on spatial partitioning strategies without strong structural constraints fails to guarantee physically accurate geometry. \textit{When scaling up to the 2.5\,km trajectory, our method's advantage becomes decisive, achieving the best overall performance (23.14\,dB PSNR and 6.62\,m D-L1).} It effectively prevents the structural degradation that plagues baselines under sparse-overlap conditions, proving that optimizing image similarity alone does not yield simulation-ready 3D structures. For completeness, we provide a more comprehensive evaluation across all six cities in the supplementary material (\autoref{tab:app_city_eval}). Furthermore, we observe that feed-forward priors suffer from compounding long-horizon drift. Attempting to substitute initialization with these drifting priors causes the optimization to collapse, dropping the PSNR to 13.40\,dB. This emphasizes that current feed-forward models cannot yet replace globally consistent pose optimization for unconstrained, city-scale reconstruction.

\begin{table*}[t]
  \centering
  \scriptsize
  \renewcommand{\arraystretch}{1.10}
  \setlength{\tabcolsep}{2.2pt}
  \caption{\textbf{Quantitative comparison with baselines on two scene scales.}
  Ann Arbor-Small (0.5k timestamps) and Atlanta-Long (5k timestamps). Best and second-best results in each column are highlighted in green and light green, respectively.}
  \label{tab:baseline_comparison}
  \resizebox{\textwidth}{!}{%
  \begin{tabular}{@{}lcccccccc@{}}
    \toprule
    & \multicolumn{4}{c}{\textbf{Ann Arbor-0.5k (0.25km)}} & \multicolumn{4}{c}{\textbf{Atlanta-5k (2.5km)}} \\
    \cmidrule(lr){2-5} \cmidrule(lr){6-9}
    \textbf{Method} & \textbf{PSNR}$\uparrow$ & \textbf{SSIM}$\uparrow$ & \textbf{LPIPS}$\downarrow$ & \textbf{D-L1}$\downarrow$
                   & \textbf{PSNR}$\uparrow$ & \textbf{SSIM}$\uparrow$ & \textbf{LPIPS}$\downarrow$ & \textbf{D-L1}$\downarrow$ \\
    \midrule
    3DGS~\cite{kerbl3Dgaussians} &
      20.00 & 0.829 & 0.581 & 32.704 &
      19.39 & 0.715 & 0.492 & 15.447 \\
    H-3DGS~\cite{kerbl2024hierarchical} &
      \second{27.59} & 0.849 & \second{0.318} & 18.450 &
      \second{21.40} & 0.747 & \second{0.380} & 14.368 \\
    CityGS~\cite{liu2024citygaussianv2} &
      24.41 & \second{0.856} & 0.409 & \second{17.053} &
      21.27 & \second{0.762} & 0.536 & \second{8.349} \\
    VGGT-Long~\cite{deng2025vggtlong} &
      -- & -- & -- & 29.632 &
      -- & -- & -- & 19.720 \\
    VGGT-Long+CityGS & 20.98 & 0.663 & 0.535 & 22.695 & 13.40 & 0.666 & 0.745 & 20.934 \\
    \textbf{Ours} &
      \best{29.99} & \best{0.917} & \best{0.240} & \best{15.158} & 
      \best{23.14} & \best{0.799} & \second{0.477} & \best{6.622} \\
    \bottomrule
  \end{tabular}%
  }
  \vspace{-1.5em}
\end{table*}

\begin{greenbox}
\textit{Partitioning and hierarchy are useful for scaling to large scenes, but they do not by themselves resolve the geometric ambiguity of narrow, long street-view trajectories. Reliable city-scale reconstruction requires structural constraints and pose consistency in addition to memory-efficient scene decomposition.}
\end{greenbox}

\begin{figure}[t]
    \centering
    \setlength{\fboxsep}{0pt}
    \begin{minipage}{0.49\linewidth}
        \centering
        \includegraphics[width=\linewidth]{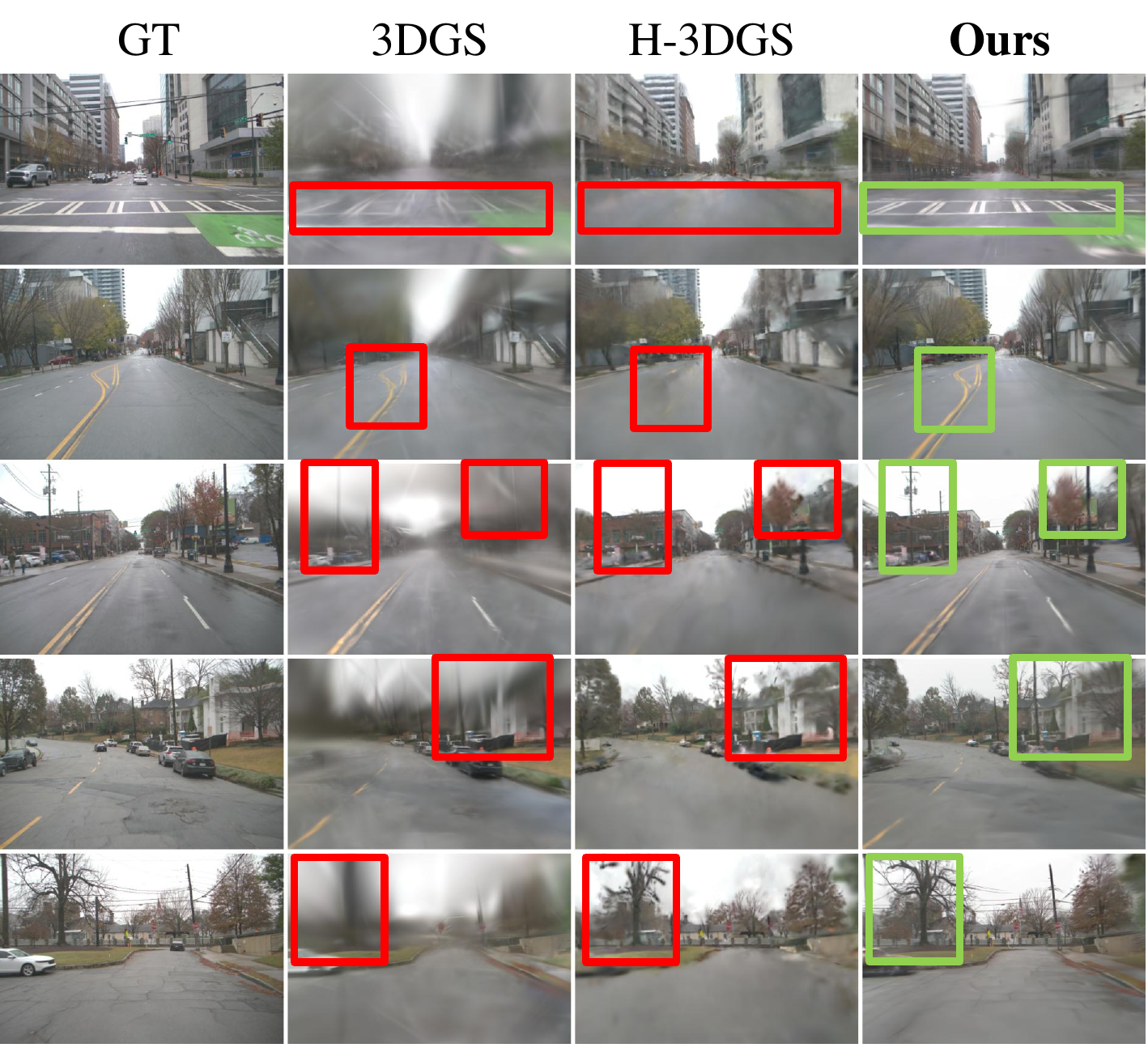}
        \vspace{2pt}
        {\small (a) Evaluation views sampled \emph{within} the training trajectory.}
    \end{minipage}
    \hfill
    \begin{minipage}{0.49\linewidth}
        \centering
        \includegraphics[width=\linewidth]{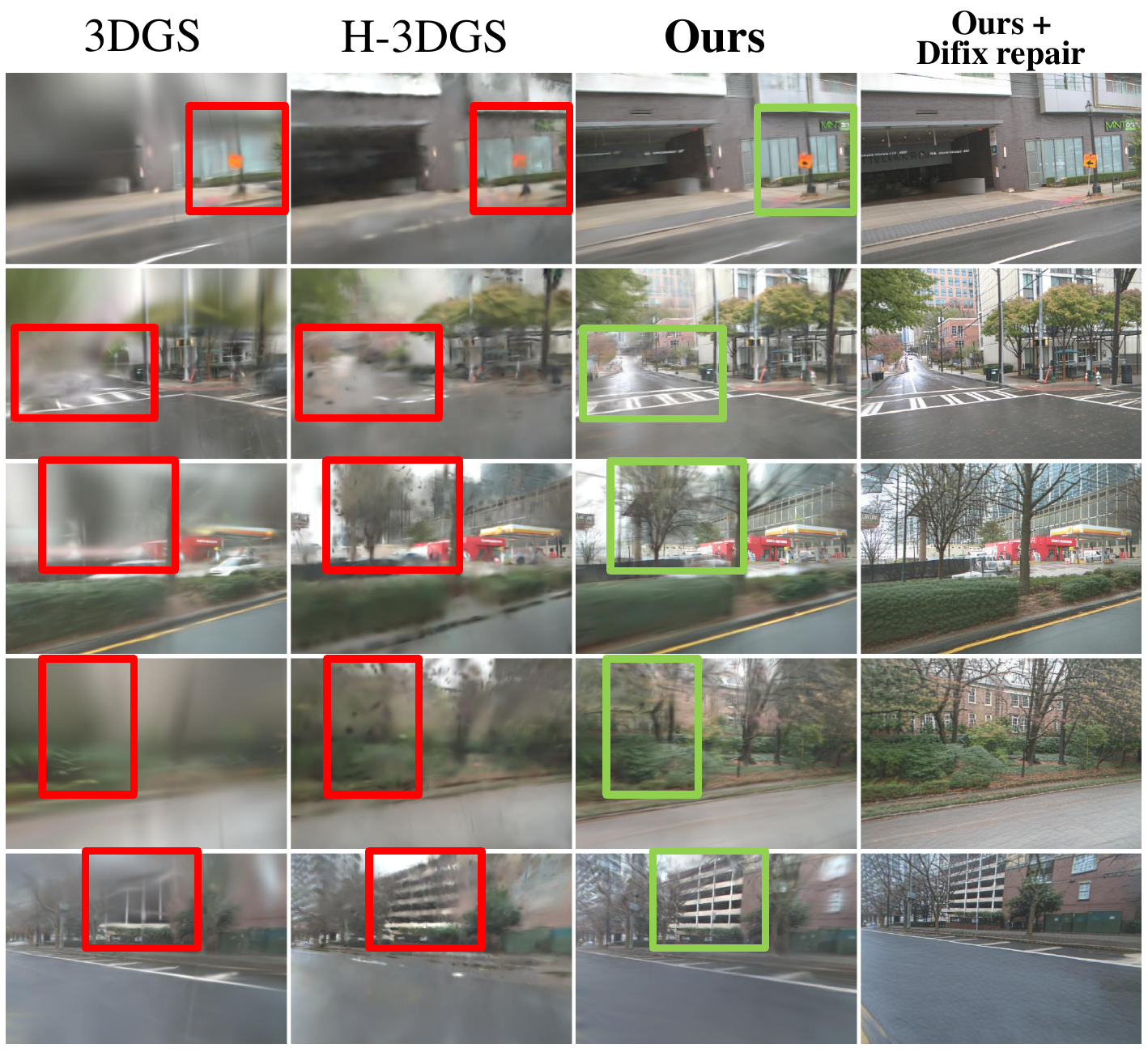}
        \vspace{2pt}
        {\small (b) Extrapolation views sampled \emph{outside} the trajectory (generalization).}
    \end{minipage}
    \caption{\textbf{Qualitative comparison with baselines.}
    We compare renderings on in-trajectory test views and extrapolated views sampled at 0.5\,m intervals outside the trajectory. For extrapolated views, we additionally present the results refined by diffusion-based repair module~\cite{wu2025difix3d+}, demonstrating its effectiveness in further mitigating artifacts and hallucinating coherent structures in unobserved regions.}
    \label{fig:baseline_qualitative}
    \vspace{-1em}
\end{figure}

\subsection{Qualitative Results}
\label{sec:qualitative_results}
\autoref{fig:baseline_qualitative} compares renderings on in-trajectory test views and extrapolated views outside the training trajectory. 
(1) On in-trajectory views, classic 3DGS often over-smooths large street scenes and loses thin structures (e.g., poles and distant façades). 
Scalable baselines recover richer content but frequently exhibit degraded texture fidelity on ground surfaces and slender objects, often accompanied by geometric inconsistency. 
\textit{Our method produces sharper textures and cleaner edges in these regions, reducing local artifacts while maintaining overall visual quality.}
(2) On extrapolated views, all methods degrade, but baselines show noticeably stronger blur and floaters, leading to unstable structures. 
\textit{In contrast, our results retain clearer scene layout with fewer floating artifacts, suggesting improved scene geometry and view robustness for simulation.}

\begin{redbox}
\textit{For closed-loop simulation, in-trajectory rendering quality is not sufficient: agents must query views beyond the recorded path. Simulation-ready urban digital twins therefore require stable geometry and coherent appearance under off-trajectory viewpoints.}
\end{redbox}

%% file: sec/6_discussion.tex
\section{Discussion}

\subsection{Performance Scalability}
\begin{figure}[t]
    \centering
    \includegraphics[width=0.99\textwidth]{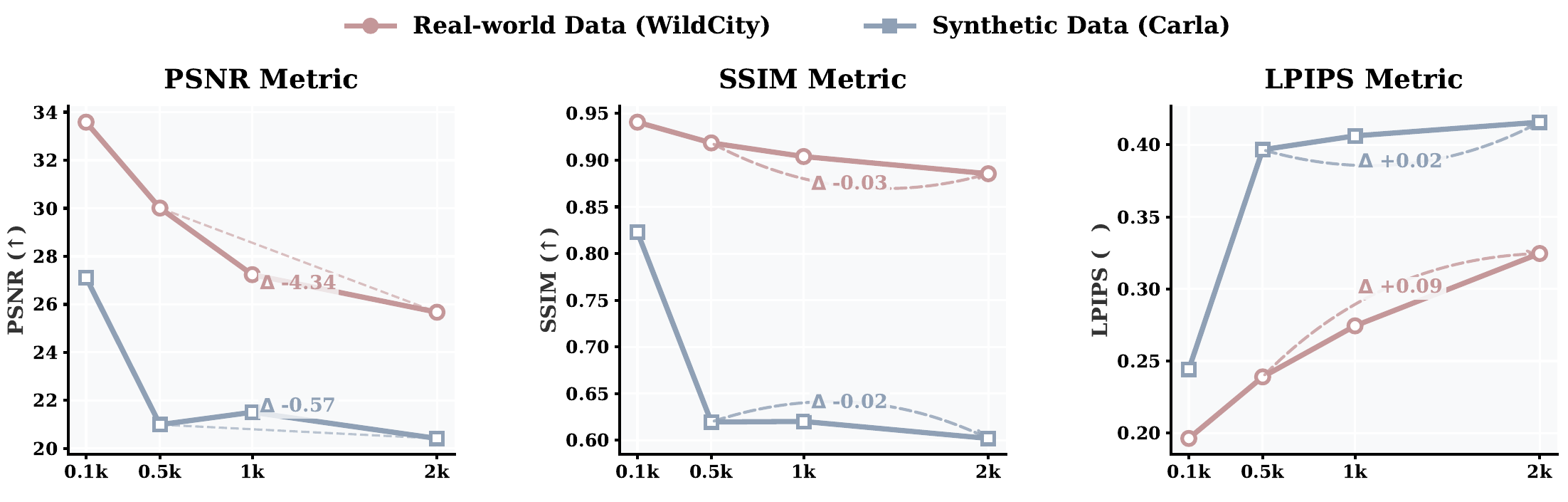}
    \setlength{\fboxsep}{0pt}
    \vspace{-3mm}
    \caption{\textbf{Quantitative evaluation across data scales.} 
    With matched Gaussian budgets at each scale, real-world \textit{WildCity} degrades more severely than synthetic Carla from 0.5k to 2k keyframes, indicating that city-scale reconstruction is constrained not only by capacity but also by accumulated pose, appearance, and uncertainty challenges.}
    \label{fig:scale}
    \vspace{-3mm}
\end{figure}

Performance scalability asks whether reconstruction quality can be maintained as scene scale increases. As shown in \autoref{fig:scale}, rendering quality degrades steadily as trajectory length grows, while peak VRAM usage and the number of optimized Gaussians increase accordingly. This indicates that current methods do not scale gracefully to continuous city-scale scenes: longer routes not only require more memory and model capacity, but also accumulate more appearance variation and pose inconsistency over space. The comparison with synthetic data further suggests that this degradation is substantially stronger in real-world settings.

\begin{table}[t]
\small
\centering
\caption{\textbf{Component ablation.} We evaluate sky modeling, rig pose optimization, and ground regularization on rendering and geometry. Notably, 2D image metrics can misalign with geometric quality: some components may slightly hurt 2D metrics yet markedly improve structural consistency and simulation usefulness (see \autoref{fig:qualitative_comparison}).}
\label{tab:ablation}
\begin{adjustbox}{width=0.99\columnwidth, center}
\setlength{\tabcolsep}{3pt}
\begin{tabular}{@{} lcc | ccc | cccc @{}}
\toprule
\multicolumn{3}{c}{\textbf{Components}} & \multicolumn{3}{c}{\textbf{Arlington (0.1km)}} & \multicolumn{3}{c}{\textbf{Atlanta (0.25km)}} \\ 
\cmidrule(lr){1-3} \cmidrule(lr){4-6} \cmidrule(lr){7-10}
\textbf{Ground} & \textbf{Sky} & \textbf{Pose Opt.} & \textbf{PSNR}$\uparrow$ & \textbf{SSIM}$\uparrow$ & \textbf{LPIPS}$\downarrow$ & \textbf{PSNR}$\uparrow$ & \textbf{SSIM}$\uparrow$ & \textbf{LPIPS}$\downarrow$ & \textbf{D-L1}$\downarrow$ \\
\midrule
\ding{55} & \ding{51} & \ding{51} & \textbf{30.14} & \textbf{0.924} & \textbf{0.186} & \textbf{26.35} & \textbf{0.864} & \textbf{0.288} & \textbf{13.11} \\
\ding{51} & \ding{55} & \ding{51} & 28.69 & 0.903 & 0.230 & 24.72 & 0.829 & 0.362 & 148.54 \\
\ding{51} & \ding{51} & \ding{55} & 29.33 & 0.913 & 0.199 & 25.51 & 0.843 & 0.320 & 14.29 \\
\midrule
\rowcolor[gray]{0.95}
\ding{51} & \ding{51} & \ding{51} & 29.60 & 0.910 & 0.217 & 25.42 & 0.815 & 0.342 & 13.26 \\ 
\bottomrule
\end{tabular}
\end{adjustbox}
\end{table}

\begin{figure}[t]
    \centering
    \includegraphics[width=0.99\textwidth]{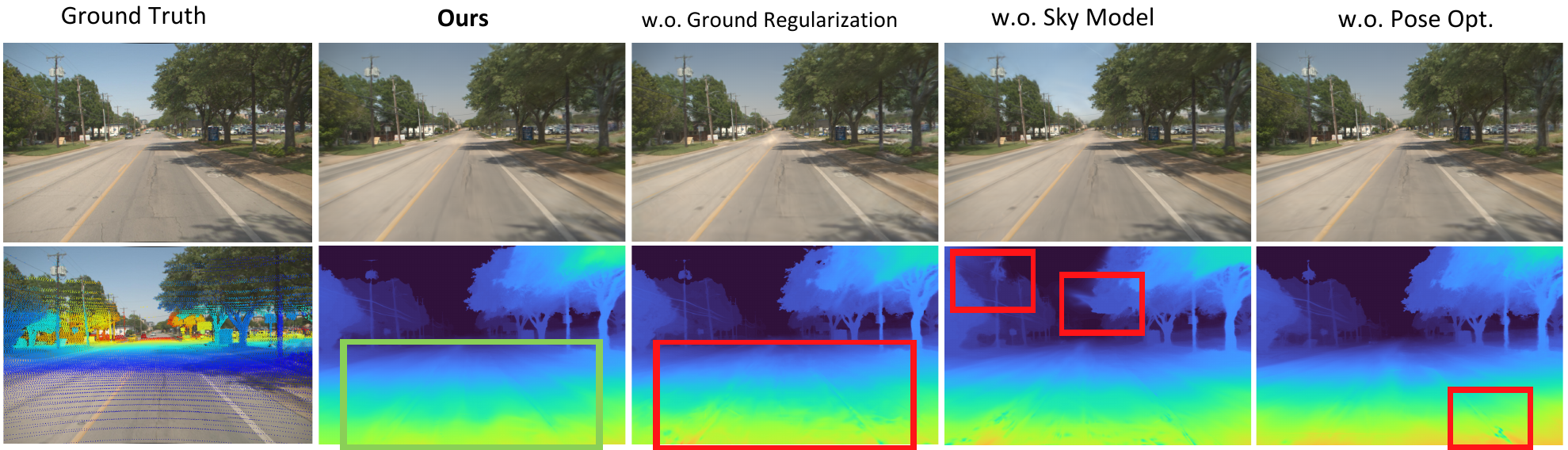}
    \setlength{\fboxsep}{0pt}
    \caption{\textbf{Qualitative results of different components.} This ablation highlights each module’s role under real-world uncertainty: removing ground regularization destabilizes road geometry, removing the sky model introduces background artifacts, and removing rig pose optimization causes multi-view misalignment from pose noise.}
    \label{fig:qualitative_comparison}
    \vspace{-3mm}
\end{figure}
\vspace{-3mm}

\subsection{View Extrapolation}
View extrapolation measures whether a model can generalize to unobserved regions under sparse or out-of-distribution viewpoints. This is especially important for interactive city digital twins, where agents must move beyond the exact recorded trajectory. As shown in \autoref{fig:baseline_qualitative}, current baselines often fail under extrapolated views, producing broken geometry, floaters, and unstable depth. Our Difix3D+~\cite{wu2025difix3d+} based post-repair improves visual quality and partially fixes these artifacts, but it remains a costly post-hoc solution and can hallucinate content when the base reconstruction is weak. This suggests that extrapolation needs to be addressed by stronger geometric priors within the reconstruction model itself.

\subsection{Data Uncertainty}
Data uncertainty arises from uncontrolled factors in real-world urban driving, including dynamic objects, illumination changes, and pose noise. The ablations in \autoref{tab:ablation} and qualitative comparisons in \autoref{fig:qualitative_comparison} confirm that these uncertainties substantially impact reconstruction fidelity and geometric consistency. While our baseline mitigates several dominant sources, \eg, a sky model for unbounded background ambiguity, rig-aware pose optimization for noisy localization, and ground regularization for weakly constrained road surfaces. The remaining errors highlight that handling real-world uncertainty remains a fundamental challenge for city-scale street-view reconstruction.

\begin{bluebox}
\textit{City-scale reconstruction remains limited by three coupled factors: incomplete 3D priors, long-horizon pose inconsistency, and weak extrapolation under sparse observations. Future evaluation should therefore move beyond 2D fidelity toward geometric correctness and simulation utility.}
\end{bluebox}

%% file: sec/7_conclusion.tex
\section{Conclusion}

We presented \textit{WildCity}, a real-world city-scale dataset for street-view reconstruction and beyond, built from long-horizon surround-view RGB-LiDAR observations collected in unconstrained urban environments. On top of this dataset, we established an urban-tailored reconstruction baseline and enabled closed-loop interaction in the reconstructed city digital twin. Together, these components provide a practical testbed for studying real-world city-scale reconstruction under realistic noise, uncertainty, and long-range spatial extent.

\noindent \textbf{Limitations.} \textit{WildCity} still has limitations. First, our semantic masks are automatically generated; although they achieve 91.58\% mIoU on 100 manually annotated images across four cities, boundary errors and rare category mistakes may remain. Second, our GPS-anchored poses avoid unbounded horizontal drift, but residual pose error persists, especially vertically: loop-closure analysis shows sub-centimeter horizontal drift and centimeter-level vertical drift per kilometer. Improving long-horizon urban pose alignment remains important future work.

\noindent \textbf{Beyond rendering.} While this paper focuses on explicit reconstruction and simulation, the broader significance of \textit{WildCity} lies in its potential to support implicit spatial understanding over long spatial and temporal horizons. Tasks such as long-form video understanding, memory, localization, and planning continue to be limited by the lack of realistic large-scale urban data. We hope \textit{WildCity} will serve not only as a benchmark for reconstruction, but also as a foundation for future research on city-scale spatial intelligence.

%% file: sec/X_suppl.tex
\clearpage
\appendix

\section{Implementation Details}
\label{sec:implementation_details}
\subsection{Data Processing}
\label{app:data_processing}

\noindent \textbf{Timestamp alignment.}
The raw sensors are asynchronous and operate at different frequencies (e.g., cameras at 10\,Hz, LiDAR at 10\,Hz, IMU at 100\,Hz, and GPS/localization at 2\,Hz), so we do not resample all streams to a common clock. Instead, we define each released keyframe by the timestamp of the front camera image,
\begin{equation}
t_k \;=\; t^{\mathrm{front}}_k ,
\end{equation}
and associate the remaining sensor measurements using their original timestamps. For each sensor stream \(s\), we select the temporally closest valid measurement
\begin{equation}
\hat{x}^{(s)}_k
\;=\;
x^{(s)}_{\,\arg\min_i \, |t^{(s)}_i - t_k|}.
\end{equation}
The ego pose is obtained by interpolation from the localization stream,
\begin{equation}
\mathbf{T}_{\mathrm{ego}}(t_k)
\;=\;
\mathrm{Interp}\!\left(\{(t^{\mathrm{loc}}_i,\mathbf{T}^{\mathrm{loc}}_i)\},\, t_k\right),
\end{equation}
where \(\mathbf{T}_{\mathrm{ego}}(t_k)\in SE(3)\) denotes the ego pose at time \(t_k\). In practice, the timestamp mismatch between the reference image and the associated LiDAR/pose measurements is typically below 50\,ms, preserving fine-grained cross-modal consistency without introducing artificial resampling artifacts.

\noindent \textbf{Motion compensation.}
Each LiDAR sweep is acquired over a finite time interval rather than instantaneously, which introduces geometric distortion when the vehicle moves during scanning. To reduce this effect, we apply sweep-level motion compensation before reconstruction. Let \(\mathbf{p}_j\) be a LiDAR point measured at firing time \(\tau_j\) within the sweep associated with keyframe \(t_k\). After transforming the point to the ego frame, we warp it to the reference time \(t_k\) via
\begin{equation}
\tilde{\mathbf{p}}_j
\;=\;
\mathbf{T}_{\mathrm{ego}}(t_k)^{-1}\,
\mathbf{T}_{\mathrm{ego}}(\tau_j)\,
\mathbf{p}_j .
\end{equation}
All LiDAR points in the sweep are deskewed in this manner and expressed at the common reference time \(t_k\). This substantially improves temporal alignment between LiDAR geometry and camera observations, leading to cleaner projected depth and more reliable supervision and evaluation.

\noindent \textbf{Mask pipeline.}
We provide semantic masks for ground, sky, and potentially movable objects. The initial masks are generated using SAM3~\cite{carion2025sam} with text prompts, and are further refined using onboard 3D tracking cuboids. To isolate truly dynamic content, we use the velocity estimates from the tracking system and classify an instance \(n\) as moving only if
\begin{equation}
\|\mathbf{v}_n\|_2 > 1 \;\text{m/s}.
\end{equation}
The final dynamic mask is formed by projecting only those tracked cuboids satisfying this criterion:
\begin{equation}
\mathcal{M}_{\mathrm{dyn}}
\;=\;
\bigcup_{n:\,\|\mathbf{v}_n\|_2 > 1}
\Pi(\mathcal{B}_n),
\end{equation}
where \(\mathcal{B}_n\) denotes the 3D cuboid of instance \(n\) and \(\Pi(\cdot)\) its image projection. Tracked instances below this threshold, such as parked vehicles or standing pedestrians, are excluded from the dynamic mask. This design separates static structure from dynamic foregrounds more reliably, making the released masks more suitable for reconstruction, evaluation, and downstream simulation.

\noindent \textbf{Release protocol.}
To make long-horizon city logs tractable for reconstruction and benchmarking, we spatially subsample keyframes at a fixed interval of 0.5\,m. Let \(\mathbf{x}_k\) denote the ego translation of keyframe \(k\). We retain keyframes such that consecutive released frames satisfy
\begin{equation}
\|\mathbf{x}_{k_{m+1}} - \mathbf{x}_{k_m}\|_2 \ge 0.5 \;\text{m}.
\end{equation}
Each route is then partitioned into contiguous 5\,km chunks according to cumulative traveled distance,
\begin{equation}
s_k
\;=\;
\sum_{j=1}^{k-1}
\|\mathbf{x}_{j+1}-\mathbf{x}_j\|_2,
\qquad
c_k
\;=\;
\left\lfloor \frac{s_k}{5000} \right\rfloor .
\end{equation}
On top of these chunks, we further release sub-trajectories of multiple lengths
\begin{equation}
L \in \{50\,\mathrm{m},\,250\,\mathrm{m},\,500\,\mathrm{m},\,1\,\mathrm{km},\,2.5\,\mathrm{km},\,5\,\mathrm{km}\},
\end{equation}
so that methods can be evaluated under progressively more challenging spatial scales.

\begin{table}[t]
\centering
\caption{\textbf{Main hyper-parameters used in our methods.} We report the parameters used for Miami 5k-keyframes data as an example. The parameters may change due to the data scale varies, such as training steps, max points and densification steps, but other optimization parameters keep the same.}
\label{tab:app_ours_hparams}
\begin{tabular}{lc}
\toprule
\textbf{Hyper-parameter} & \textbf{Value} \\
\midrule
Training steps & 600,000 \\
Maximum number of Gaussians & 30,000,000 \\
Initial mean learning rate & $2\times10^{-3}$ \\
Final mean LR multiplier & $10^{-2}$ \\
Depth loss weight $\lambda_{\mathrm{depth}}$ & $10^{-1}$ \\
Depth mode & disparity \\
Pose optimization start step & 1 \\
Pose optimization learning rate & $10^{-3}$ \\
Pose optimization regularization & $10^{-5}$ \\
Pose optimization mode & rig \\
Densification start step & 10,000 \\
Densification stop step & 60,000 \\
Densification interval & 100 \\
Densify portion & 0.005 \\
Ground optimization steps & 10,000 \\
Spherical harmonics degree & 1 \\
Sky appearance embedding dim & 16 \\
Sky MLP layers & 3 \\
Sky MLP hidden width & 64 \\
Sky skip connection & layer 1 \\
\bottomrule
\end{tabular}
\end{table}

\subsection{Our Method Implementation}
\label{app:method_impl}
Our method is built on top of a 3DGS training pipeline with several urban-specific modifications. Unless otherwise specified, all experiments share the same default optimization settings across scenes, while scene-dependent adjustments are limited to the total training steps and the Gaussian capacity budget. The main hyper-parameters of our method are summarized in \autoref{tab:app_ours_hparams}.

\noindent \textbf{Sky model.}
We model the sky separately using a lightweight view-dependent MLP rather than 3D Gaussians. Given a viewing direction $\mathbf{d}\in\mathbb{R}^3$, we first apply a sinusoidal encoder with degree range $[0,6)$, and then concatenate the encoded direction with a learnable per-image appearance embedding of dimension 16. The resulting feature is passed through a 3-layer MLP with hidden width 64, output dimension 3, and a skip connection at the second layer. Formally, for image $i$, the sky color is predicted as
\[
\mathbf{c}_{\mathrm{sky}}
=
f_{\mathrm{sky}}\!\left([\gamma(\mathbf{d}),\,\mathbf{e}_i]\right),
\]
where $\gamma(\cdot)$ is the directional encoder, $\mathbf{e}_i\in\mathbb{R}^{16}$ is the appearance embedding, and $f_{\mathrm{sky}}$ is the MLP. A sigmoid activation is applied to the output RGB values. At test time, when image indices are unavailable, we replace $\mathbf{e}_i$ with the mean appearance embedding over the training set.

\noindent \textbf{Rig pose optimization.}
We optimize the ego pose at each keyframe together with the camera extrinsics relative to the ego frame, rather than treating each camera pose independently. Pose optimization is enabled from the beginning of training (\texttt{pose\_opt\_start}=1) in rigid-rig mode, with learning rate $10^{-3}$ and regularization weight $10^{-5}$. This rigid factorization is particularly important for surround-view data, where small localization errors can otherwise accumulate into multi-view inconsistency over long trajectories.

\noindent \textbf{Ground regularization.}
For Gaussians identified as ground, we apply additional priors to stabilize road geometry, including slice-wise distortion regularization on ground height, shortest-axis alignment, and opacity regularization. These losses are activated only during the early stage of training using a curriculum schedule of 20k steps (\texttt{ground\_curriculum\_steps}=10000), when the geometry is most unstable.

\noindent \textbf{Depth supervision.}
We apply external depth supervision throughout training using a disparity-based loss with weight $10^{-1}$ (\texttt{depth\_mode=disparity} and \texttt{depth\_lambda}=0.1). This encourages better geometric consistency in large urban scenes, especially in texture-poor regions.

\noindent \textbf{Gaussian optimization and densification.}
We train for 600k steps with an initial mean learning rate of $2\times10^{-3}$ and a final learning-rate multiplier of $10^{-2}$. The Gaussian capacity is capped at 30M primitives. Densification begins at 10k iterations, ends at 60k iterations, is performed every 100 steps, and adds $0.5\%$ of the current primitive count per refinement step (\texttt{densify\_portion}=0.005). We use spherical harmonics of degree 1 (\texttt{sh\_degree}=1).

\noindent \textbf{Multi-GPU training.}
For long sequences, we follow the distributed design of GrendelGS~\cite{zhao2024scaling} and shard Gaussian parameters across multiple GPUs. In our default setting, training is performed on two H200 GPUs. This keeps the per-GPU memory footprint bounded and enables training at city scale.

\noindent \textbf{Training protocol.}
All experiments are trained with dynamic masks enabled and sky modeling activated. Checkpoints are saved at 200k and 400k iterations, and evaluation is performed periodically throughout training, with denser evaluation in the earlier optimization stage and final reporting at 400k steps. We do not use the bilateral grid option in the reported setting.

\subsection{Baseline Implementation}
\label{app:baseline_impl}
We implement all baselines using their official or publicly released codebases whenever available, and adapt only the data loading and preprocessing interfaces necessary to support the WildCity format. All methods use the same camera intrinsics, poses, and train/validation split, and are evaluated under the same masking-aware protocol used in the main paper.

\noindent \textbf{3DGS.}
We use the standard 3DGS training pipeline~\cite{kerbl3Dgaussians} as a classic point of reference. Since vanilla 3DGS is not designed for multi-kilometer scenes, it mainly serves as a lower-bound baseline for large-scale street-view reconstruction.

\noindent \textbf{H-3DGS.}
For H-3DGS~\cite{kerbl2024hierarchical}, we follow the hierarchical Gaussian training procedure and keep its scene decomposition and merging strategy unchanged. This baseline evaluates whether hierarchical Gaussian representations alone are sufficient for continuous street-view trajectories.

\noindent \textbf{CityGS.}
For CityGaussianV2~\cite{liu2024citygaussianv2}, we follow the released blockwise training and large-scene rendering setup. This baseline is particularly relevant because it is designed for large-scale Gaussian reconstruction through spatial partitioning.

\noindent \textbf{VGGT-Long and VGGT-Long+CityGS.}
We use VGGT-Long~\cite{deng2025vggtlong} as a feed-forward long-sequence geometry prior. Since it does not directly produce photorealistic renderings in our evaluation protocol, we report its depth accuracy after aligning its reconstruction to our metric point-cloud frame via a single global similarity transform. For VGGT-Long+CityGS, we replace the COLMAP initialization in CityGS with the aligned VGGT-Long reconstruction and retain the rest of the CityGS training pipeline unchanged. We also tested VGGT-X as a feed-forward initializer, but on both multi-thousand-view sequences it produced only extremely sparse and unstable point clouds, so we exclude it from the main quantitative table.

\noindent \textbf{Training budget and stopping criterion.}
We do not enforce a fixed iteration budget across all baselines, because their optimization structures differ substantially. Vanilla 3DGS uses a single-stage end-to-end optimization, H-3DGS uses a base stage followed by post-optimization, and CityGS-based methods use a multi-stage pipeline consisting of coarse optimization, spatial partitioning, block-wise trimming, merging, and final evaluation. In our experiments, 3DGS is trained for 50k iterations on Ann Arbor-0.5k and 400k iterations on Atlanta-5k, with all urban-specific components disabled, including sky modeling, depth supervision, dynamic masking, and curriculum strategies. H-3DGS follows its released hierarchical pipeline with 50k base iterations plus 15k post-optimization iterations on Ann Arbor-0.5k, and 30k base iterations plus 15k post-optimization iterations on Atlanta-5k. For CityGS and VGGT-Long+CityGS, we convert LiDAR depth to inverse-depth supervision and run the released CityGaussianV2 pipeline. On Ann Arbor-0.5k, both methods use 30k coarse iterations followed by 10k block-wise trim iterations after validation performance saturates. On Atlanta-5k, both methods use 90k coarse iterations followed by 90k block-wise trim iterations, which was the strongest stable budget we completed for the city-scale setting.

For CityGS-based methods, the trim stage uses the released 4$\times$4 spatial partition with block-wise parallel optimization, and we preserve the same downstream optimization schedule when replacing the default COLMAP initialization with the aligned VGGT-Long reconstruction. All runs are conducted on comparable high-memory datacenter GPUs in the H100/H200 class rather than under an artificially fixed device budget. We therefore report each baseline under a strong and stable configuration within its native training paradigm.

\section{Additional Experiments}
\label{app:additional_experiments}

\subsection{Evaluation Across Cities}
\label{app:across_cities}
To study how reconstruction performance varies across geography, we extend our evaluation to additional cities beyond the main sequences used in the paper. This experiment serves two purposes. First, it measures whether our method can generalize across cities with different street scenes and styles. Second, it reveals how city morphology affects city-scale reconstruction, especially the urban scenes, like Atlanta and rurual scenes, like Grand Rapids.

We consider representative cities with distinct urban characteristics, such as dense downtown regions, residential grids, broad arterial roads, and high-speed corridors. For each city, we evaluate the same method under the same protocol and summarize the results in \autoref{tab:app_city_eval}. We also include a cross-city generalization setting, where initialization and hyper-parameters are all the same, in order to test the robustness of the pipeline across different urban morphologies.

The results shows that cities with denser intersections, more severe occlusions, and more heterogeneous street layouts pose greater challenges than cities with more regular road structure and cleaner visibility. These results complement the main paper by showing that the difficulty of city-scale reconstruction depends not only on route length, but also on the morphology and operational complexity of the urban environment.

\begin{table*}[t]
\centering
\caption{\textbf{Quantitative comparison across different cities (2.5km).} We report PSNR, SSIM, and LPIPS on 2.5km sequences. Performance varies across cities due to different urban morphology and scene complexity; for example, Atlanta contains highly textured street scenes and frequent dynamic objects, making reconstruction more challenging. Despite this variation, most cities achieve PSNR above 25\,dB, indicating that the reconstruction pipeline remains robust across diverse real-world environments.}
\label{tab:app_city_eval}
\begin{tabular*}{\textwidth}{@{\extracolsep{\fill}}lcccc}
\toprule
\textbf{City (2.5km)} & \textbf{PSNR $\uparrow$} & \textbf{SSIM $\uparrow$} & \textbf{LPIPS $\downarrow$} & \textbf{D-L1 $\downarrow$}\\
\midrule
Atlanta   & 23.59 & 0.8157 & 0.3648 & 10.5906 \\
Arlington & 24.95 & 0.8321 & 0.3421 & 14.6566 \\
Ann Abour & 25.33 & 0.9040 & 0.2937 & 24.1174 \\
Eden Prairie     & 26.56 & 0.8771 & 0.3155 & 16.9166 \\
Grand Rapids      & 26.22 & 0.9165 & 0.2250 & 17.0085 \\
Miami     & 25.07 & 0.8403 & 0.3216 & 14.9335 \\
\bottomrule
\end{tabular*}
\end{table*}

\begin{figure}[t]
    \centering
    \includegraphics[width=1.00\textwidth]{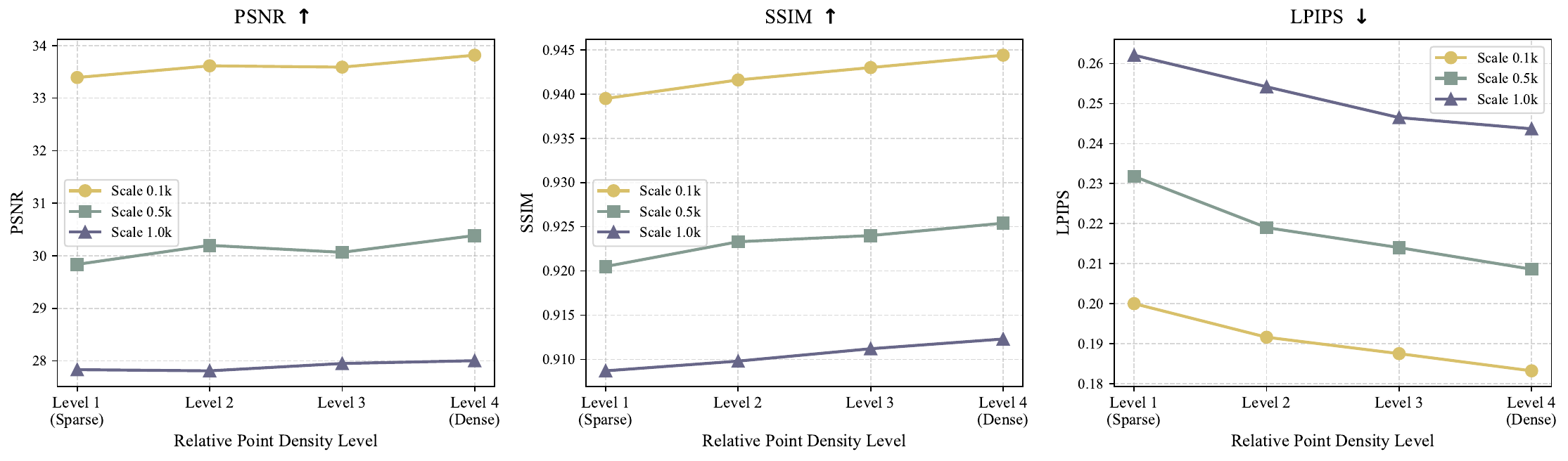}
    \caption{\textbf{Quantitative comparison on impact of Gaussian primitive numbers across different scales.} Density Levels 1 to 4 correspond to point counts of {3M, 4M, 5M, 6M} for Scale 0.1k; {7M, 9M, 11M, 13M} for Scale 0.5k; and {15M, 20M, 25M, 30M} for Scale 1.0k, respectively.}
    \label{impact}
\end{figure}

\subsection{Effect of the Number of Gaussian Primitives}
\label{app:gaussian_count}
We further study the trade-off between reconstruction quality and computational cost under different Gaussian budgets. Specifically, we vary the maximum number of Gaussian primitives while keeping the rest of the training setup fixed, and report the resulting rendering quality, depth accuracy, peak GPU memory, and training time. As illustrated in \autoref{impact}, a higher relative point density level yields an upward trend in the PSNR of the reconstructed scene, alongside a steady increase in SSIM and a steady decrease in LPIPS. This demonstrates that regardless of the scale, adding more Gaussian points enhances the rendering of scene details, thereby improving the overall quantitative metrics.

This experiment helps clarify whether better quality primarily comes from improved modeling design or from simply allocating more Gaussian capacity. It also reveals the practical cost of scaling current pipelines to large urban scenes. In general, increasing the number of primitives improves quality up to a point, but also raises memory usage and optimization cost substantially. This trade-off is particularly important for city-scale logs, where Gaussian growth can quickly become a bottleneck.



\subsection{Extrapolation Offset Study}
\label{app:extrapolation_offset}
\begin{figure*}[t]
\centering
\includegraphics[width=0.95\textwidth]{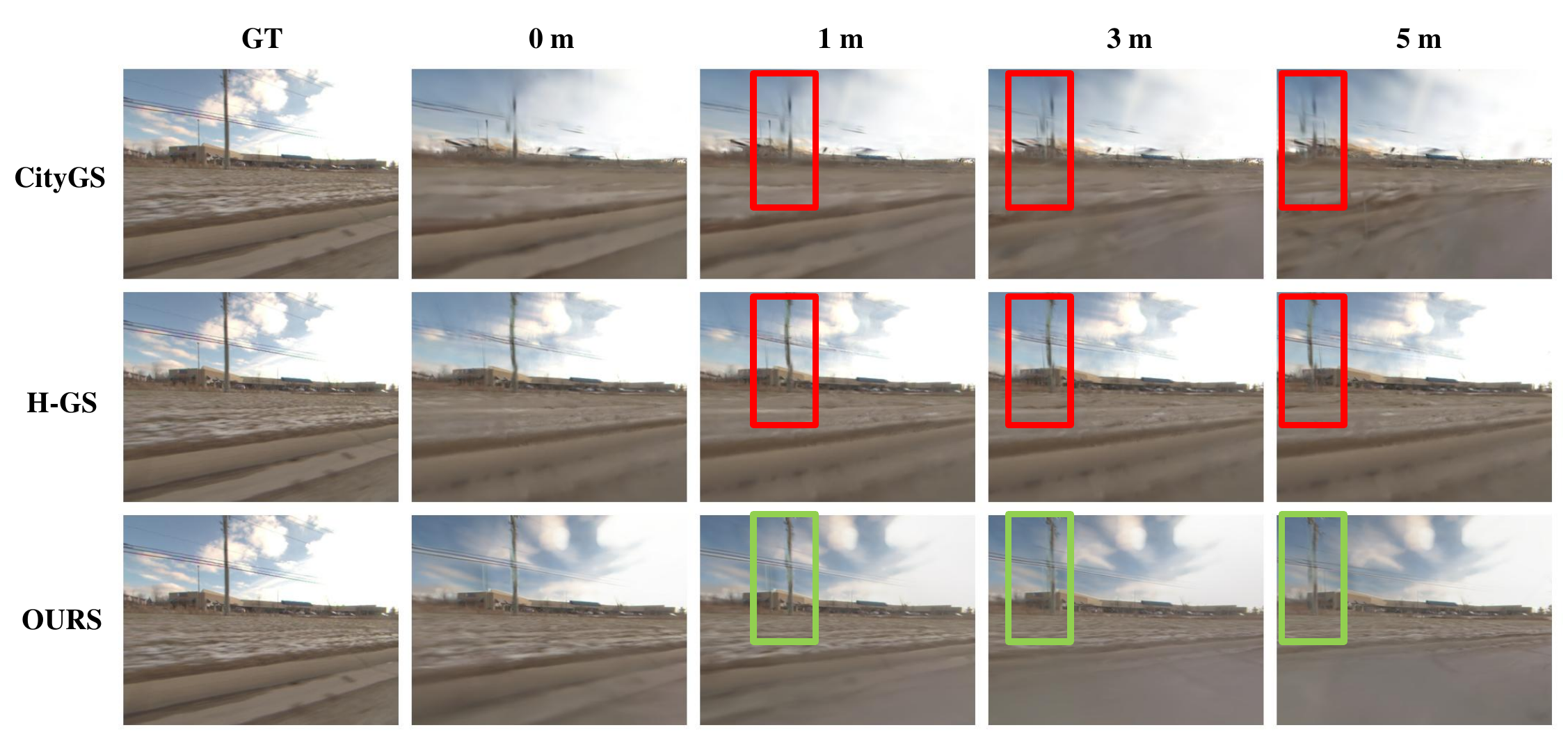}
\caption{\textbf{Qualitative extrapolation under increasing off-trajectory offsets.}
We render viewpoints with lateral offsets of 1\,m, 3\,m, and 5\,m from the recorded trajectory. This figure shows representative anchor views and compares the recorded-view reference with reconstructions from our method, CityGS, and H-3DGS.}
\label{fig:app_extrapolation_offsets_1}
\end{figure*}

\begin{figure*}[t]
\centering
\includegraphics[width=0.99\textwidth]{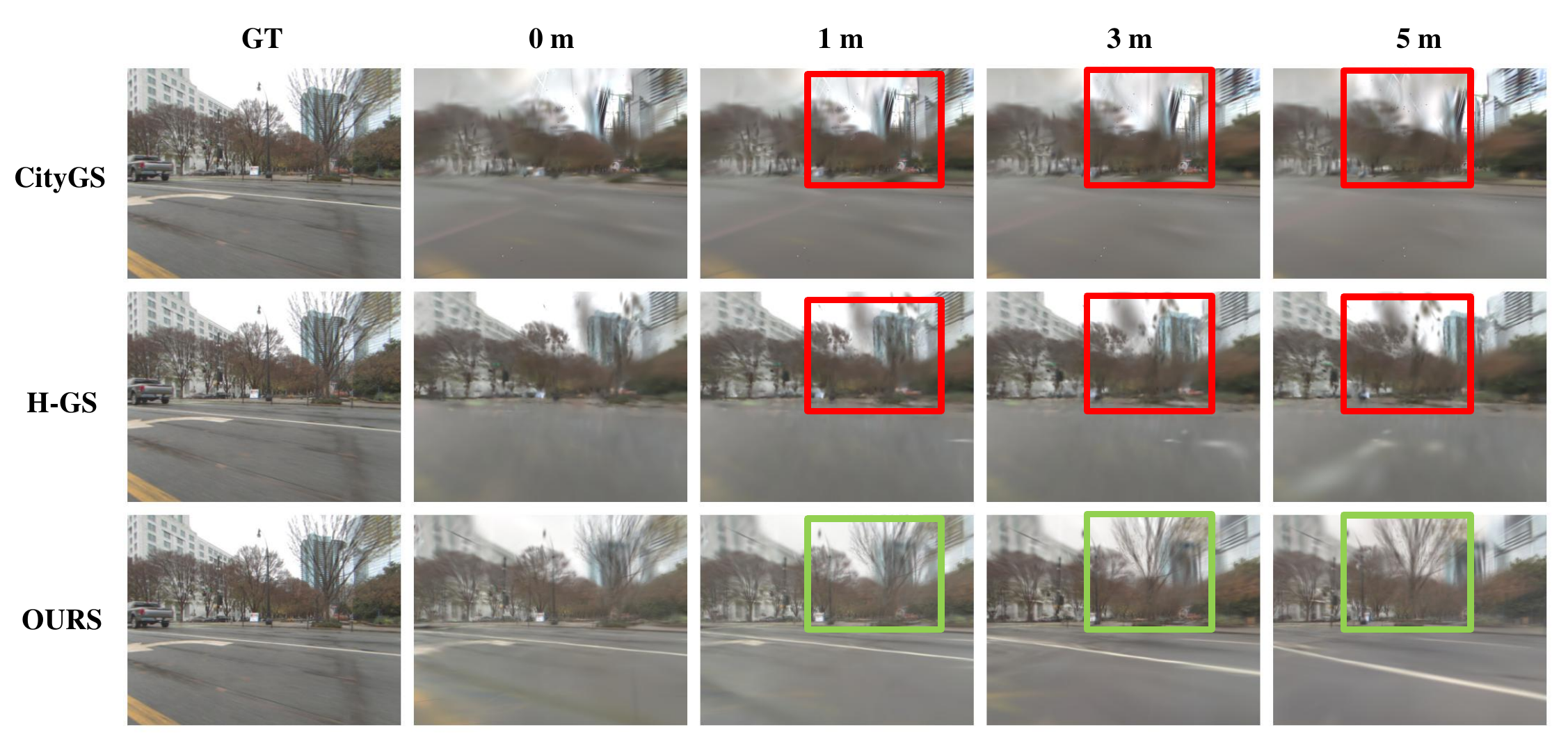}
\caption{\textbf{Additional qualitative extrapolation results.}
Continued examples under 1\,m, 3\,m, and 5\,m off-trajectory offsets.}
\label{fig:app_extrapolation_offsets_2}
\end{figure*}

\begin{figure*}[t]
\centering
\includegraphics[width=0.99\textwidth]{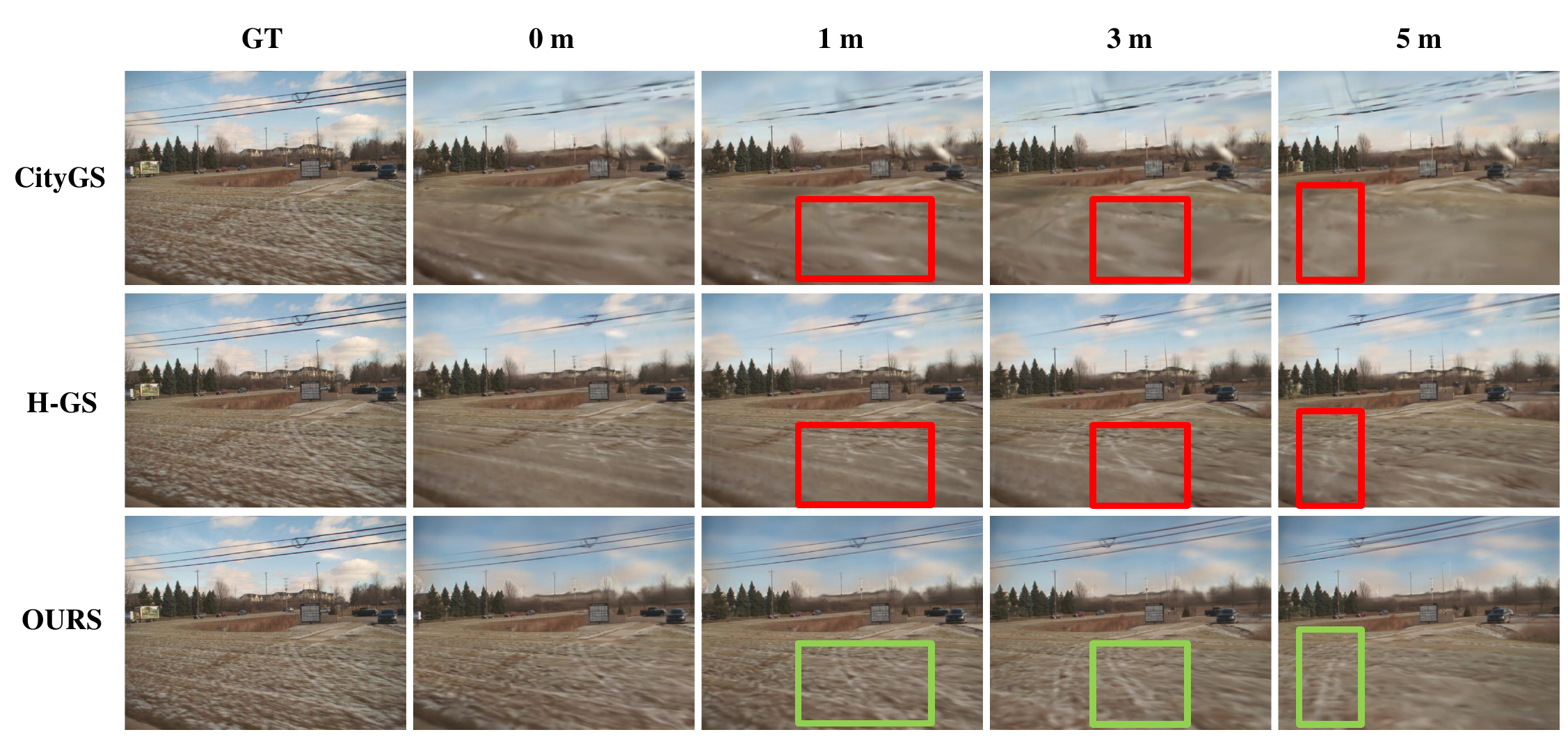}
\caption{\textbf{Additional qualitative extrapolation results.}
Continued examples under 1\,m, 3\,m, and 5\,m off-trajectory offsets.}
\label{fig:app_extrapolation_offsets_3}
\end{figure*}

\begin{figure*}[t]
\centering
\includegraphics[width=0.99\textwidth]{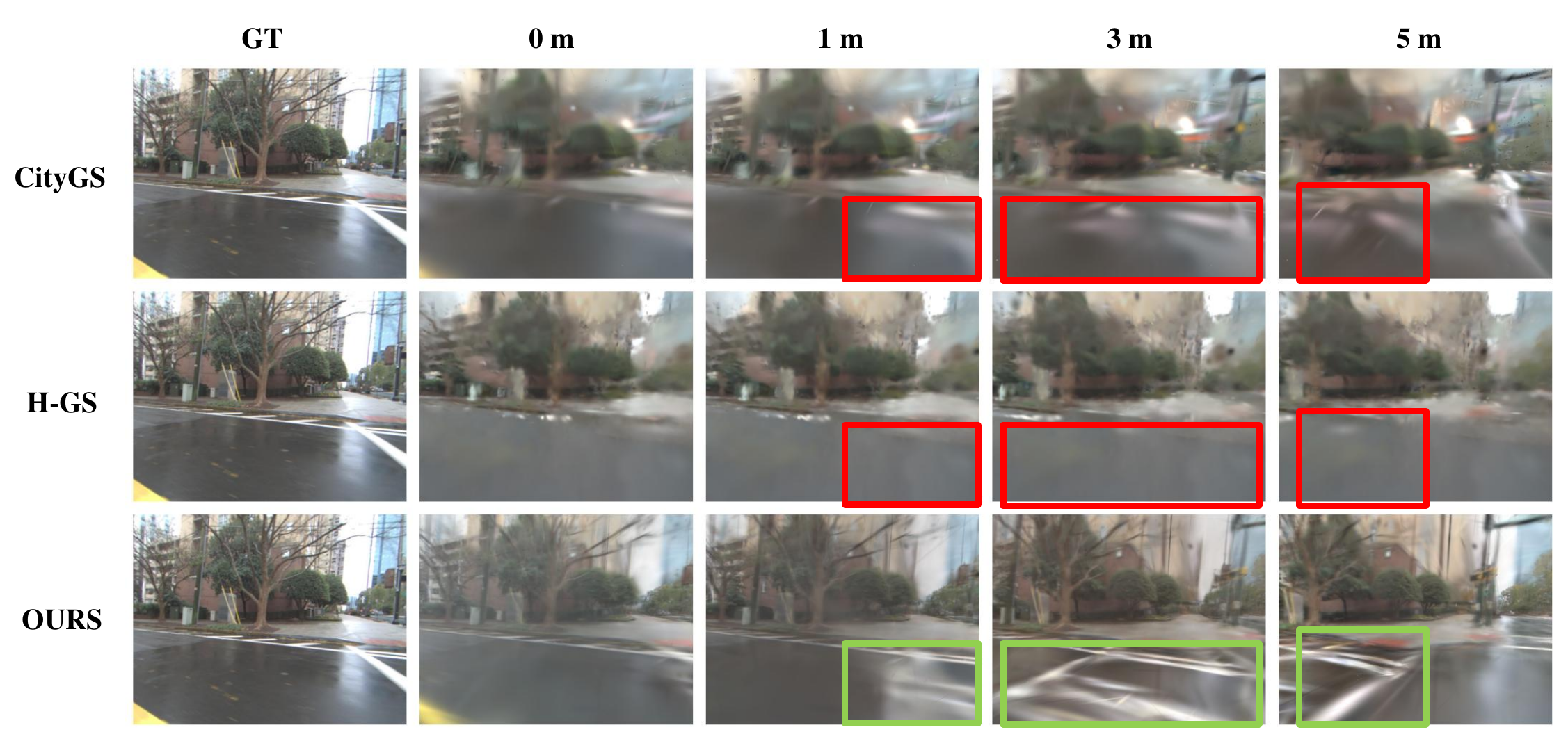}
\caption{\textbf{Additional qualitative extrapolation results.}
Continued examples under 1\,m, 3\,m, and 5\,m off-trajectory offsets.}
\label{fig:app_extrapolation_offsets_4}
\end{figure*}

\begin{figure*}[t]
\centering
\includegraphics[width=0.99\textwidth]{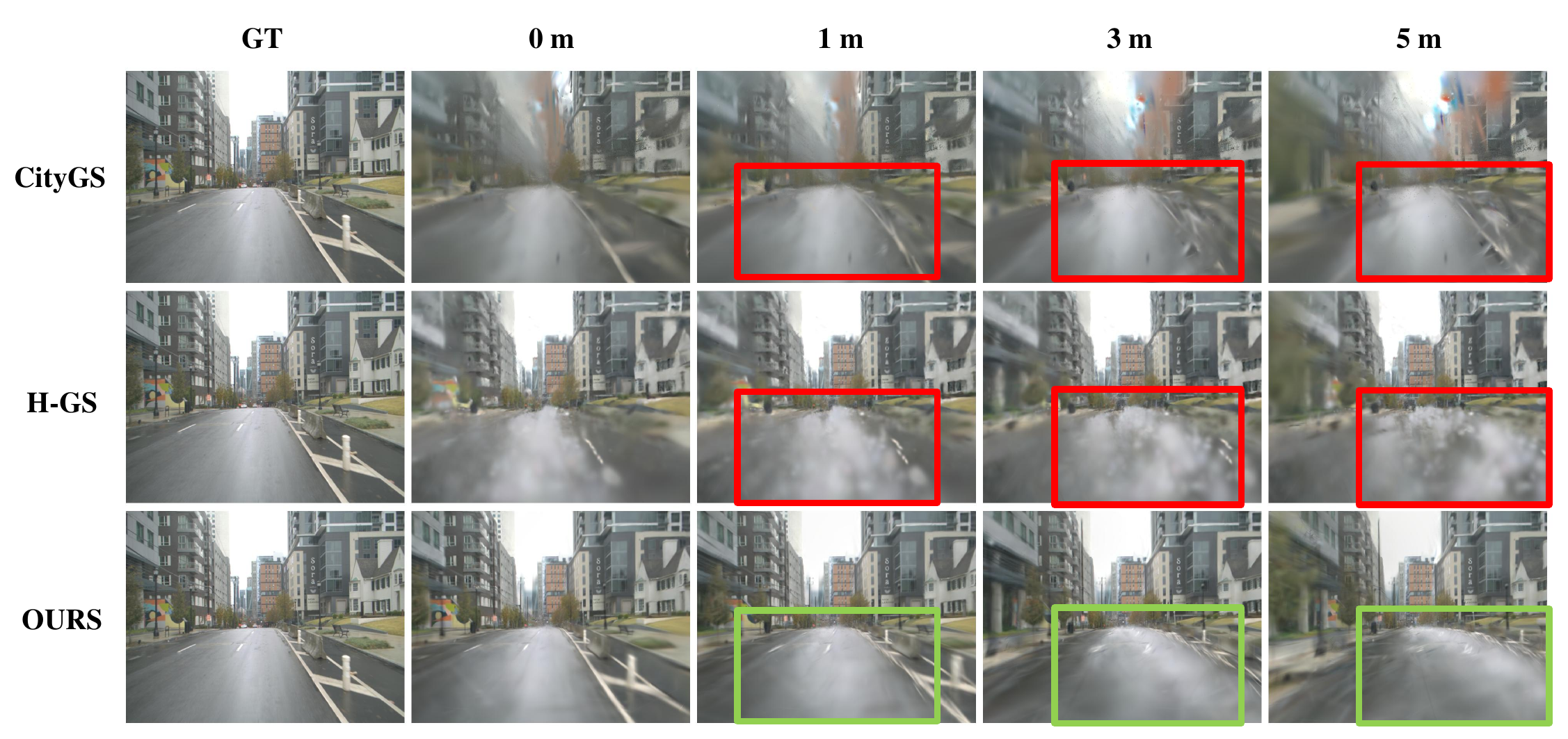}
\caption{\textbf{Additional qualitative extrapolation results.}
Continued examples under 1\,m, 3\,m, and 5\,m off-trajectory offsets.}
\label{fig:app_extrapolation_offsets_5}
\end{figure*}

To further analyze view extrapolation, we study off-trajectory viewpoints with increasing lateral offsets from the recorded trajectory. Compared with the milder offsets considered in the initial draft, we use a more challenging setting with offsets of $[1, 3, 5]$\,m, which makes the degradation trend more visually explicit on long street-view trajectories. Qualitative visualizations are shown in \autoref{fig:app_extrapolation_offsets_1}--\autoref{fig:app_extrapolation_offsets_5}. Each row contains five representative anchor views and compares the recorded-view reference together with the renderings produced by our method, CityGS~\cite{liu2024citygaussianv2}, and H-3DGS~\cite{kerbl2024hierarchical} as the camera progressively departs from the original trajectory.

This experiment complements the qualitative comparison in the main paper by making the extrapolation difficulty more explicit. Across all examples, the visual quality of every method degrades as the viewpoint moves from the recorded pose to 1\,m, 3\,m, and 5\,m off-trajectory, but the failure modes differ substantially. CityGS~\cite{liu2024citygaussianv2} and H-3DGS~\cite{kerbl2024hierarchical} increasingly exhibit blurred structures, broken ground surfaces, unstable sky boundaries, and floating artifacts as the offset grows. In contrast, our method remains visibly more stable in both ground and sky regions, which we attribute to the additional geometric supervision imposed on these two structurally important components. The advantage becomes especially clear at larger offsets, where the new viewpoints depart further from the original sampling manifold and therefore stress global geometric consistency rather than local photometric fitting alone.

